\pdfoutput=1
\documentclass[journal]{IEEEtran}
 
\usepackage{amsmath,amssymb,amsfonts}
\usepackage{bm}
\usepackage{algorithm}
\usepackage{algpseudocode}
\usepackage{graphicx}
\usepackage{subcaption}
\usepackage{xcolor}
\definecolor{linkblue}{RGB}{0, 0, 150}
\usepackage[
    colorlinks=true,
    urlcolor=linkblue,
    citecolor=black,
    linkcolor=black
]{hyperref}
\usepackage{enumitem}
\usepackage{booktabs}

\newcommand{\lamT}{\bm{\lambda}_T} 
\newcommand{\sigT}{\bm{\sigma}_T}  
\newcommand{\R}{\mathbb{R}}

\newcommand{\norm}[1]{\left\lVert#1\right\rVert}
\newcommand{\Kmu}{\mathcal{K}_\mu}
\newcommand{\Kmus}{\mathcal{K}^*_\mu}

\newcommand{\Mlam}{\mathcal{M}_{\lambda}}
\newcommand{\Msig}{\mathcal{M}_{\sigma}}

\newcommand{\Frho}{F_\rho}
\newcommand{\sigN}{\sigma_N}             
\newcommand{\lamN}{\lambda_N}            
\newcommand{\unit}{\hat{\bm{\sigma}}_T}

\newcommand{\Jc}{J_c}               
\newcommand{\Mmat}{M}               

\begin{document}

\title{DiffContact: Differentiable Contact Dynamics \\ for Policy Learning and Contact-Implicit \\ Trajectory Optimization}

\author{Gijeong~Kim$^{1}$, Seungbin~Oh$^{1}$, Dongyun~Kang$^{1}$, and~Hae-Won~Park$^{1}$%
\thanks{$^{1}$The authors are with the Department of Mechanical Engineering, Korea Advanced Institute of Science and Technology, Daejeon 34141, South Korea.}}

\maketitle
 
\begin{abstract}
This paper proposes DiffContact, a smooth analytic gradient for frictional contact, and applies it to
policy learning and contact-implicit trajectory optimization for legged
locomotion. %
The exact contact sensitivity is available in closed form but describes changes within the current contact mode, providing no first-order information toward neighboring contact modes. 
The proposed gradient replaces
this sensitivity only in the backward pass, leaving the hard-contact forward 
solve unchanged and the realized rollout feasible. A central-path relaxation of the
de~Saxc\'e-corrected second-order-cone complementarity problem for Coulomb
friction smooths the conditions defining the separating, clamping, and sliding
modes. The relaxed sensitivity is evaluated at a point on the central path
obtained by a Newton iteration initialized from the hard-contact solution. The
sensitivity remains continuous through contact-mode transitions and converges
to the exact sensitivity as the relaxation vanishes. Using 100,000 samples and
less than 2~min of training on a desktop PC, policy learning with the proposed
gradient produces a policy that transfers to hardware without
fine-tuning. On hardware, the robot under a trotting policy withstands pushes and traverses debris and
slippery ground. In trajectory optimization, the proposed gradient reaches contact-rich solutions missed by exact and Signorini-relaxed gradients. Project page: \url{https://diffcontact.github.io}
\end{abstract}


\section{Introduction}\label{sec:intro}

\IEEEPARstart{T}{he} first-order gradient through contact in a differentiable
physics simulator provides a common computational primitive for two classes of
methods in contact-rich robot control. The first is policy learning, in which model-free
reinforcement learning is widely used and policy-gradient estimates are
computed from rollouts across thousands of parallel
environments~\cite{hwangbo2019learning,rudin2022learning,
makoviychuk2021isaac}. First-order policy
learning~\cite{Heess2015,degrave2019differentiable} preserves the same objective
but replaces the score-function (REINFORCE)
estimate~\cite{Williams1992,Sutton1999} with a typically lower-variance
first-order estimate~\cite{Mohamed2020}. This estimate can be obtained by differentiating through the simulator, with comparable policies learned from fewer samples~\cite{Xu2022}.
The second is trajectory optimization. Contact-implicit trajectory
optimization (CITO) methods determine when and where contacts are made and
broken without a prescribed contact-mode
sequence~\cite{Mordatch2012,Posa2014,Manchester2019}. Shooting-based DDP/iLQR
methods~\cite{Tassa2012,Neunert2018} use the same transition
Jacobians as the policy-learning estimator and have been applied online in
receding-horizon control for contact-rich locomotion~\cite{Kim2023}.

We study both tools in legged locomotion, where separating, clamping, and sliding contacts recur within every stride, and both are limited by the contact sensitivity provided by differentiable simulation. 
A simulator may be differentiable without providing a useful first-order direction toward a neighboring contact
mode~\cite{Suh2022,Freeman2021}. The exact contact sensitivity is available in
closed form~\cite{Werling2021,LeLidec2025}, but it is evaluated with the active
set, or contact-mode assignment, held fixed and therefore describes only the
response within the current mode. A clamped contact is linearized as remaining
clamped, providing no first-order information toward liftoff or slip, and the
sensitivity changes discontinuously when the active set changes. In
contact-implicit optimization, this mode-restricted sensitivity keeps the local
search near the contact sequence of the nominal
trajectory~\cite{Kim2023, Werling2021, Pang2023}. In first-order policy learning, the same
Jacobians enter backpropagation through time (BPTT), where they can produce
fragile or ineffective optimization in contact-rich
systems~\cite{Suh2022,Freeman2021,Metz2021,Georgiev2024}. 

Prior work addresses contact sensitivity in policy learning and CITO in different ways. In policy learning, methods shorten the differentiation horizon~\cite{Xu2022,Georgiev2024}, use a reduced-order model for the backward pass~\cite{Song2024}, or smooth contact in the forward rollout~\cite{Schwarke2024}. Shorter horizons can help with long-horizon gradient instability, but do not change the contact sensitivity within each window. In CITO, transcription methods optimize contact impulses alongside the trajectory~\cite{Posa2014,Manchester2019,Dietz2026}, whereas shooting methods rely on transition Jacobians from the contact dynamics~\cite{Tassa2012,Neunert2018,Kim2023}. Other approaches smooth the forward contact model~\cite{Kurtz2025,Pang2023}, differentiate relaxed contact solutions~\cite{Howell2022,LeCleach2024}, or estimate gradients across perturbed solves~\cite{Suh2022rs,LeLidec2024rs}. For shooting-based CITO with a hard-contact forward solve, however, the exact contact sensitivity remains tied to the current contact mode and offers limited first-order guidance toward a different mode~\cite{Werling2021}. Relaxing only the Signorini condition during differentiation~\cite{Kim2023} partly addresses this limitation by providing directions for contact make/break transitions, but leaves sliding and normal–tangential coupling unaddressed.

We therefore propose DiffContact, a contact sensitivity obtained by relaxing the full frictional contact condition only in the backward pass, providing search directions toward neighboring contact modes while retaining the hard-contact forward solve. The de Saxcé correction expresses frictional contact as a single second-order-cone complementarity condition~\cite{DeSaxce1998,LeLidec2024}. Relaxing this condition along a central path~\cite{Acary2026} couples normal and tangential contact sensitivities across separation, clamping, and sliding. These coupled sensitivities are available in every contact mode, allowing the optimizer to change modes through gradient-based updates. We further investigate their role in guiding transitions to sliding, where a normal-only relaxation lacks the required tangential information.

We test the effect of this sensitivity in two optimal-control settings using a quadruped robot: first-order policy learning and contact-implicit trajectory optimization. In each setting, we compare the exact~\cite{LeLidec2025}, Signorini-relaxed~\cite{Kim2023}, and proposed sensitivities while keeping the forward contact model, objective, and initialization fixed. The policy-learning experiments ask whether single-environment BPTT can produce locomotion policies that transfer directly to hardware. The trajectory-optimization experiments ask whether the proposed sensitivity improves contact-rich motion generation and recovery from unmodeled external pulls by enabling contact-mode changes and foot repositioning.

\par\vspace{3pt}
The main contributions are:
\begin{itemize}[leftmargin=*]
\item DiffContact, an analytic backward contact sensitivity that relaxes the complete de~Saxc\'e-corrected frictional contact law along its central path while retaining the hard-contact forward solve. It provides first-order directions toward separation, sliding, and clamping in every contact mode and recovers the exact sensitivity as the relaxation vanishes.

\item The proposed gradient in BPTT-based first-order policy learning and contact-implicit trajectory optimization guides contact-mode changes and yields improved policies and lower-cost contact-rich trajectories.%

\item Hardware demonstration of quadruped policy learning with the proposed
gradient for trotting, pacing, and gait-free locomotion. Trotting policies obtained
using 100,000 samples in under 2~min are tested on hardware
without fine-tuning under pushes, debris, and low-friction surfaces.
\end{itemize}
\vspace{3pt}

The remainder of this paper is structured as follows.
Section~\ref{sec:related} reviews related work and Section~\ref{sec:bg} the
optimal control and contact dynamics background. Section~\ref{sec:problem}
states the required contact sensitivity, Section~\ref{sec:method} derives the
proposed gradient, and Section~\ref{sec:impl} the implementation details.
Sections~\ref{sec:apg} and~\ref{sec:cito} present the policy-learning and
trajectory-optimization demonstrations, and Section~\ref{sec:conclusion}
concludes.

\section{Related Work}\label{sec:related}

\subsection{Contact simulation and differentiable contact}\label{sec:related_contact}
\emph{Forward models.} Rigid frictional contact is commonly modeled using three coupled principles~\cite{LeLidec2024,Hwangbo2018}: the Signorini condition, Coulomb's friction-cone bound, and Moreau's maximum-dissipation principle (MDP)~\cite{Moreau1988}. Their combination yields a nonsmooth nonlinear complementarity problem for the contact impulse, solved at each time step using iterative methods such as projected Gauss--Seidel and staggered projections~\cite{LeLidec2024}. Linear-complementarity (LCP) models polygonalize the friction cone, allowing the use of pivoting methods~\cite{AnitescuPotra1997}. Convex cone-complementarity (CCP) models relax strict Signorini complementarity to obtain a convex contact problem. During sliding, this can permit a nonzero normal force together with separating normal velocity, leading to artificial normal motion and, in time-stepping simulations, force across a gap~\cite{LeLidec2024}. MuJoCo employs a related convex soft-contact formulation~\cite{todorov2012mujoco}. Compliant models use finite contact stiffness and dissipation; with suitable regularization, convex formulations can also handle near-rigid contact~\cite{Castro2022}.

In this work, the hard-contact forward model in CITO retains the three principles of the rigid nonlinear complementarity problem. We use per-contact iteration~\cite{Hwangbo2018}, solving sliding-contact subproblems by bisection with the de~Saxc\'e correction~\cite{LeLidec2024MDP}. For BPTT policy learning, we instead use RaiSim~\cite{Hwangbo2018} for efficient forward rollouts because training requires repeated simulations at a fine time step. This also allows us to compare BPTT with PPO in the same simulation environment, although standard RaiSim can deviate from maximum dissipation during sliding~\cite{LeLidec2024MDP}. In both settings, we derive the contact gradient from the de Saxcé representation of the rigid frictional contact law~\cite{DeSaxce1998,Acary2018}.

\emph{Differentiating contact.} Finite differences estimate contact sensitivities through repeated evaluations with perturbed inputs, with the number of evaluations growing with the input dimension~\cite{todorov2012mujoco,Tassa2012}. Near contact-mode transitions, the estimates can depend on the perturbation size. Automatic differentiation instead applies the chain rule to the operations executed by the simulator~\cite{hu2020difftaichi,Freeman2021,degrave2019differentiable}. Analytic approaches can combine derivatives of rigid-body algorithms ~\cite{Featherstone2008,carpentier2018analytical} with implicit differentiation of the contact equations at the converged contact solution~\cite{Werling2021,LeLidec2025,BelbutePeres2018}. The proposed method follows this analytic approach, but evaluates sensitivity at a relaxed central-path point of the frictional contact condition.

\subsection{Differentiable simulation for policy learning}\label{sec:related_learning}
Model-free reinforcement learning (RL) is widely used for legged locomotion. Score-function policy-gradient methods do not require derivatives of the simulator~\cite{Schulman2017}, and large-scale parallel simulation accelerates the collection of large sample batches for stable policy updates~\cite{rudin2022learning}. This approach has produced hardware locomotion~\cite{hwangbo2019learning,Tan2018,Lee2020,Miki2022} and example-guided motor skills~\cite{Peng2018}. First-order policy learning instead differentiates the rollout dynamics to estimate the policy gradient~\cite{degrave2019differentiable,Freeman2021}, which can improve sample efficiency~\cite{Xu2022,Zhang2025}. Related model-based RL methods differentiate a learned world model rather than a physics simulator~\cite{Heess2015,Hafner2020}; DayDreamer applies this approach to quadruped hardware learning~\cite{Wu2022}.

First-order estimates through long unrolled dynamics can be unstable or less effective than zeroth-order estimates, particularly under stiff or nonsmooth dynamics~\cite{Suh2022,Metz2021}. To mitigate these difficulties, SHAC~\cite{Xu2022} combines first-order estimates with short-horizon actor--critic training, and AHAC~\cite{Georgiev2024} adapts the differentiation horizon to avoid stiff contact dynamics. For systems without contact transitions, differentiable flight models have supported quadrotor-policy training and hardware transfer~\cite{Zhang2025,Wiedemann2023}. Contact-rich locomotion additionally requires the backward pass to traverse repeated changes among separating, clamping, and sliding contacts.

Recent locomotion methods modify the dynamics used for differentiation in different ways to obtain informative contact gradients. Schwarke et al.~\cite{Schwarke2024} smooth contact activation with distance-dependent sigmoid scaling and differentiate the resulting simulation using automatic differentiation. This improves gradient informativeness but also modifies the forward contact dynamics. Song et al.~\cite{Song2024} retain Isaac Gym for forward simulation while computing gradients through a state-aligned single-rigid-body surrogate that relies on a prescribed gait schedule. Both approaches demonstrate hardware transfer with policies trained using 128 and 64 parallel simulation environments, respectively. The present work instead uses an articulated hard-contact simulator for the forward rollout and a backward-only relaxation of its frictional contact sensitivity, without a reduced-order surrogate or smoothed forward dynamics. Moreover, we demonstrate that a single rollout per update suffices for BPTT with the proposed gradient to learn locomotion policies that transfer directly to hardware.

\subsection{Contact-implicit trajectory optimization}\label{sec:related_cito}
Contact-implicit trajectory optimization (CITO) determines contact timing and location jointly with the state and control trajectories. Early formulations include contact-invariant optimization~\cite{Mordatch2012} and formulations that encode contact activation and separation through complementarity constraints~\cite{yunt2006trajectory,Posa2014}. The resulting mathematical program must resolve the discrete contact-mode structure embedded in these constraints.

CITO methods differ in where they handle the contact law. \emph{Direct transcription} includes contact forces or impulses among the trajectory decision variables~\cite{Posa2014}; related formulations combine centroidal dynamics with full kinematics~\cite{Dai2014}, or use orthogonal collocation~\cite{Patel2019} and variational integration~\cite{Manchester2019}. Consensus-based algorithms separate trajectory-wide dynamics constraints from per-step contact constraints and coordinate the subproblems through ADMM~\cite{Aydinoglu2022,Aydinoglu2024}. Phase-based formulations instead optimize contact timing through stance and swing durations~\cite{Winkler2018,kim2025dynamically}. \emph{Shooting}-based methods resolve contact within each simulation step and provide transition Jacobians to the trajectory optimizer~\cite{Tassa2012,kong2023hybrid, kurtz2022contact}. When computed within a fixed active set, hard-contact sensitivities can provide limited guidance for contact-mode changes~\cite{Werling2021}.

Smoothed contact sensitivities can be obtained by modifying either the contact model or its differentiation. Compliant and barrier-smoothed models replace hard contact in the planning model and may introduce force at a distance or interpenetration~\cite{Tassa2012,Neunert2018,Kurtz2025}. Local smoothing regularizes the contact map analytically or by averaging over perturbations~\cite{Pang2023,Suh2022rs,LeLidec2024rs}; sampling-based variants require multiple perturbed solves. Interior-point approaches differentiate a relaxed system at a nonzero central-path parameter~\cite{Howell2022,LeCleach2024}; the differentiated point is an intermediate iterate of the forward solver, so its relaxation level is tied to the values visited by the solver and the gradient is obtained within that solver. 
Relaxing only the Signorini condition during differentiation~\cite{Kim2023,kim2022contact} decouples the relaxed point from the forward solve, which remains hard. However, the friction cone stays unrelaxed, so a clamped contact has no tangential direction toward slip. The proposed gradient instead applies the central-path relaxation to the full de~Saxc\'e-corrected frictional cone, thereby coupling normal and tangential sensitivities across separating, clamping, and sliding contacts. Its evaluation point is not a byproduct of the forward solve but is computed by a separate Newton iteration from the hard-contact solution, so the relaxation parameter can be chosen for the backward pass alone.

\section{Background}\label{sec:bg}

Shooting-based trajectory optimization and BPTT-based policy learning use the
transition Jacobians $(f_x,f_u)$ in a backward recursion. Forming these Jacobians
through contact requires the contact-impulse sensitivity
$\partial\bm{\lambda}/\partial(A,\bm{b})$. This section reviews the two
optimal-control formulations and the forward dynamics that resolve contact, and
derives the analytic contact-impulse sensitivity for each contact mode. The second-order 
cone complementarity problem (SOCCP) formulation of
frictional contact used by the proposed gradient computation is also reviewed.

\subsection{Optimal-control formulation}\label{sec:bg_oc}
With state $\bm{x}$, control input $\bm{u}$, and discrete-time dynamics
$\bm{x}_{t+1}=f(\bm{x}_t,\bm{u}_t)$, trajectory optimization and first-order
policy learning can be posed as finite-horizon optimal-control problems of
the form,
\begin{equation}\label{eq:ocp}
\begin{aligned}
&\min_{\xi}\ \mathbb{E}_{\bm{x}_0\sim p_0}\Bigl[
\textstyle\sum_{t=0}^{H-1}\ell_t(\bm{x}_t,\bm{u}_t)
+ \ell_H(\bm{x}_H)\Bigr]\\
&\text{s.t.}\quad
\bm{x}_{t+1}=f(\bm{x}_t,\bm{u}_t),\qquad
\bm{u}_t=\bm{u}_t(\xi;\bm{x}_t),
\end{aligned}
\end{equation}
where the decision variable $\xi$ determines the controls through
$\bm{u}_t(\xi;\bm{x}_t)$, $\ell_t$ and $\ell_H$ are the running and terminal
costs, and $p_0$ is the initial-state distribution. The two settings differ
in the choice of $\xi$, $p_0$, horizon, and solver. Both rely on the
transition Jacobians
$f_x:=\partial\bm{x}_{t+1}/\partial\bm{x}_t$ and
$f_u:=\partial\bm{x}_{t+1}/\partial\bm{u}_t$.
Throughout, $d(\cdot)$ denotes the total differential with respect to an
arbitrary perturbation, $d(\cdot)/d\eta$ the corresponding total
derivative (sensitivity) with respect to a parameter $\eta$,
$\partial(\cdot)/\partial(\cdot)$ a partial derivative with the remaining
arguments held fixed, and $\nabla$ the gradient of a scalar function.

\emph{Trajectory optimization (shooting-based DDP/iLQR).} Here the decision variable is
the open-loop control sequence $\xi=\bm{u}_{0:H-1}$ from a fixed initial state
$\bm{x}_0$. Differential dynamic programming~\cite{Mayne1966,Tassa2012} is a shooting method that
solves \eqref{eq:ocp} by alternating an explicit forward rollout of $f$ with a backward
Riccati sweep derived from Bellman's principle of optimality. The sweep forms a local quadratic approximation of the state--control value $Q$, whose derivatives combine derivatives of the cost $\ell$, the transition Jacobians $(f_x,f_u)$, and the next-step value derivatives $(V_x',V_{xx}')$. The iterative linear--quadratic regulator (iLQR/iLQG) uses the Gauss--Newton approximation that omits the second-order dynamics terms~\cite{li2004iterative,Tassa2012}. Under this approximation,
\begin{equation}\label{eq:ddp_Q}
Q_x = \ell_x + f_x^\top V_x',\qquad
Q_{xx} = \ell_{xx} + f_x^\top V_{xx}' f_x,
\end{equation}
with $Q_u,Q_{uu},Q_{ux}$ formed analogously. Minimizing the local quadratic model over the control increment yields the
feedforward and feedback gains, whose back-substitution updates the cost-to-go for the
preceding step and carries the sweep back to $t=0$. Its feasibility-driven
multiple-shooting variant (FDDP)~\cite{Mastalli2020} admits an infeasible
state--control initialization and closes the dynamics gaps over the iterations,
widening the basin of attraction under poor initialization.

\emph{Policy learning (first-order, via BPTT).} Here the decision variable is the
policy parameter $\xi=\theta$ of a feedback policy $\bm{u}_t=\pi_\theta(\bm{x}_t)$,
with initial states sampled from $\bm{x}_0\sim p_0$. For a truncated rollout of length $H$, a finite-horizon loss is $L(\theta)=\mathbb{E}_{\bm{x}_0\sim p_0}\sum_{t=0}^{H-1}\ell_t$.
A learned value function can be added as the terminal term to bootstrap the
remaining return, as in actor--critic variants such as SHAC~\cite{Xu2022}. The gradient of $L$ is obtained
by backpropagation through time (BPTT)~\cite{Werbos1990} along the closed-loop rollout. With
$\pi_{x,t}:=\partial\pi_\theta(\bm{x}_t)/\partial\bm{x}_t$, the state and action
gradients follow the backward recursion as,
\begin{equation}\label{eq:bptt}
\begin{aligned}
\nabla_{\bm{x}_t}L &= \ell_{x,t}+\pi_{x,t}^\top\ell_{u,t}
+\bigl(f_x+f_u\pi_{x,t}\bigr)^{\!\top}\nabla_{\bm{x}_{t+1}}L,\\
\nabla_{\bm{u}_t}L &= \ell_{u,t}+f_u^\top\nabla_{\bm{x}_{t+1}}L.
\end{aligned}
\end{equation}
The first recursion propagates the future loss backward through the closed-loop
Jacobian $f_x+f_u\pi_{x,t}$, while the second gives the sensitivity of the rollout
loss to the action at each step. The parameter gradient then follows as
$\nabla_\theta L=\mathbb{E}_{\bm{x}_0\sim p_0}\sum_t
\bigl(\partial\pi_\theta(\bm{x}_t)/\partial\theta\bigr)^\top\nabla_{\bm{u}_t}L$. 
In practice, reverse-mode automatic differentiation accumulates this recursion over
the unrolled rollout using the transition Jacobians $f_x$ and $f_u$ supplied by the
differentiable simulator at each step~\cite{hu2020difftaichi,Freeman2021}.

\emph{Relation to reinforcement learning.} Model-free reinforcement learning (RL)
instead optimizes the expected return of a \emph{stochastic} policy and estimates its
gradient by the score-function (likelihood-ratio) rule,
$\nabla_\theta J=\mathbb{E}_{\tau\sim\pi_\theta}\!\bigl[R(\tau)\,\nabla_\theta\log
p(\tau;\theta)\bigr]$~\cite{Williams1992,Sutton1999}, with $\tau$ a sampled trajectory.
This \emph{zeroth-order} estimator requires no dynamics derivatives and can be applied
to nonsmooth or even discontinuous dynamics, with policy stochasticity providing a
randomized smoothing of the objective~\cite{Duchi2012, Suh2022}, at the cost of high
variance~\cite{Mohamed2020,Recht2019}. In contrast, the first-order gradient above is lower-variance and
sample-efficient~\cite{Heess2015,Xu2022}, but nonsmooth and near-rigid contact
can produce degenerate or ill-conditioned gradients~\cite{Suh2022,Metz2021}.

\subsection{Dynamics model and the frictional contact problem}\label{sec:bg_contact}
A legged robot interacts with the environment through intermittent
frictional contact, modeled here by contact impulses. Its state is described
by the configuration $\bm{q}\in\mathcal{Q}\cong\R^{n_q}$ and generalized
velocity $\bm{v}\in T_{\bm{q}}\mathcal{Q}\cong\R^{n_v}$, where
$T_{\bm{q}}\mathcal{Q}$ denotes the tangent space at $\bm{q}$. The dynamics
are integrated at the velocity level using a time-stepping
scheme~\cite{Moreau1988,Stewart2000} with a semi-implicit Euler step of size
$\Delta t$,
\begin{equation*}
\Mmat(\bm{q})\,(\bm{v}^+-\bm{v})
= \bigl(S^\top \bm{u} - h(\bm{q},\bm{v})\bigr)\Delta t + \Jc(\bm{q})^\top \bm{\lambda},
\end{equation*}
with symmetric positive-definite joint-space inertia $\Mmat\succ0$, nonlinear
(Coriolis, centrifugal, and gravitational) terms $h$, actuation selector $S$,
control input (joint torque) $\bm{u}$, and contact Jacobian
$\Jc:T_{\bm{q}}\mathcal{Q}\to\R^{3n_c}$ mapping generalized velocities to
the velocities of the $n_c$ contact points; $\bm{\lambda}\in\R^{3n_c}$
stacks the contact impulses over the step.

Solving for the next-step velocity $\bm{v}^+$ decomposes it into the free velocity
$\bm{v}_{\rm free}$ and a contact contribution,
\begin{equation}\label{eq:vplus}
\bm{v}^+ = \bm{v}_{\rm free} + \Mmat^{-1}\Jc^\top \bm{\lambda},
\qquad
\bm{v}_{\rm free} := \bm{v}
+ \Mmat^{-1}(S^\top\bm{u}-h)\,\Delta t .
\end{equation}
Mapping \eqref{eq:vplus} to the contact frame gives the contact-space
(Delassus) relation
\begin{equation*}
\bm{\sigma}=A\bm{\lambda}+\bm{b},\qquad
A:=\Jc\Mmat^{-1}\Jc^\top\succeq0,\qquad
\bm{b}:=\Jc\bm{v}_{\rm free},
\end{equation*}
where $\bm{\sigma}$ is the contact velocity, $A$ is the \emph{Delassus matrix} (the contact-space inverse inertia) and $\bm{b}$ is the free-contact term. To suppress contact drift, the
normal component of $\bm{b}$ is augmented by the position-level correction $\phi(\bm{q}_t)/\Delta t$, where $\phi(\bm{q}_t)$ is the signed normal gap, positive in separation. Thus, $\sigma_N=\bm{J}_N\bm{v}^+ + \phi(\bm{q}_t)/\Delta t$, where $\bm{J}_N$ denotes the normal rows of $\Jc$, and $\Delta t\,\sigma_N$ approximates the next-step signed gap.

For a single contact, dropping the index $i$ and writing
$(\bm{\lambda},\bm{\sigma})\in\R^3\times\R^3$ with Delassus pair
$(A,\bm{b})$, frictional point contact is governed by three
principles~\cite{Stewart2000,Moreau1988,LeLidec2024}. The \emph{Signorini}
condition expresses contact unilaterality as
\[
0 \le \sigN \;\perp\; \lamN \ge 0 .
\]
\emph{Coulomb's law} confines the impulse to the friction cone,
\[
\bm{\lambda}\in\Kmu :=
\{(\lamN,\lamT)\in\R\times\R^2:\ \norm{\lamT}\le\mu\lamN\},
\]
while the \emph{maximum-dissipation principle} (MDP) selects the admissible
friction impulse that maximizes dissipated power,
\[
\lamT \in
\arg\max_{\norm{\bm{\gamma}}\le\mu\lamN}
\bigl(-\sigT^\top\bm{\gamma}\bigr).
\]
For sliding contact ($\sigT\neq\bm{0}$), the MDP gives
$\lamT=-\mu\lamN\hat{\bm{\sigma}}_T$, where
$\hat{\bm{\sigma}}_T:=\sigT/\norm{\sigT}$. These conditions define the
following contact modes~\cite{LeLidec2024}:
\begin{equation}\label{eq:modes}
\begin{cases}
\bm{\lambda} = \bm{0},\ \ \sigN > 0,
& \text{(\emph{separating})},\\[2pt]
\bm{\sigma} = \bm{0},\ \ \bm{\lambda} \in \operatorname{int}\Kmu,
& \text{(\emph{clamping})},\\[2pt]
\sigN = 0,\ \ \lamT=-\mu\lamN\,\hat{\bm{\sigma}}_T,
& \text{(\emph{sliding})}.
\end{cases}
\end{equation}
The separating mode carries no contact impulse, the clamping mode corresponds
to sticking with an interior friction impulse, and the sliding mode places the
impulse on the cone boundary, $\norm{\lamT}=\mu\lamN$. The active mode is
determined by the solution $(\bm{\lambda},\bm{\sigma})$ rather than known
a priori, giving rise to the combinatorial structure of the contact problem.

\subsection{The de~Saxc\'e SOCCP and its Jordan-algebra form}\label{sec:bg_soccp}
The Signorini condition, Coulomb friction cone, and MDP can be recast as a
single conic complementarity condition using the de~Saxc\'e
correction~\cite{DeSaxce1998,Acary2018}. The correction
$\Gamma_\mu(\bm{\sigma})=(\mu\norm{\sigT},0,0)^\top$ offsets the normal
velocity by the dissipation prescribed by the MDP:
\begin{equation}\label{eq:soccp}
\Kmu \ni \bm{\lambda} \;\perp\; \bm{\sigma}+\Gamma_\mu(\bm{\sigma}) \;\in\; \Kmus ,
\qquad \Kmus=\mathcal{K}_{1/\mu}.
\end{equation}
The orthogonality ``$\perp$'' comprises three coupled conditions: primal
feasibility $\bm{\lambda}\in\Kmu$, dual feasibility
$\bm{\sigma}+\Gamma_\mu(\bm{\sigma})\in\Kmus$, and complementarity
$\bm{\lambda}^\top(\bm{\sigma}+\Gamma_\mu(\bm{\sigma}))=0$.
This complementarity condition holds for the three contact modes in
\eqref{eq:modes}~\cite{LeLidec2024}. For separating and
clamping contacts, it follows from $\bm{\lambda}=\bm{0}$ and
$\bm{\sigma}=\bm{0}$, respectively. For sliding contact, $\sigN=0$ and the MDP relation
$\lamT=-\mu\lamN\hat{\bm{\sigma}}_T$ give
$\lamT^\top\sigT=-\mu\lamN\norm{\sigT}$. This term cancels the contribution
$\mu\lamN\norm{\sigT}$ from the de~Saxc\'e correction, so that
$\bm{\lambda}^\top(\bm{\sigma}+\Gamma_\mu(\bm{\sigma}))=0$.

To put this complementarity in the form used by standard second-order cone
solvers, the interior-point treatment of friction~\cite{Acary2026} rewrites it
through the Jordan algebra of a self-dual cone. This requires a self-dual
(symmetric) cone $\mathcal{K}=\mathcal{K}^{*}$, and the friction cone is
self-dual only at unit friction; we therefore apply the standard unit-friction
rescaling~\cite{Acary2026}
\begin{equation}\label{eq:scaling}
\tilde{\bm{\lambda}} := (\mu\lambda_N;\bm{\lambda}_T),
\qquad
\tilde{\bm{\sigma}} :=
\bigl(\sigma_N/\mu+\norm{\bm{\sigma}_T};\bm{\sigma}_T\bigr).
\end{equation}
This maps $\bm{\lambda}\in\Kmu$ and
$\bm{\sigma}+\Gamma_\mu(\bm{\sigma})\in\Kmus$ onto the self-dual
second-order cone
$\mathcal{K}=\{(w_0,\bar{\bm{w}})\in\R\times\R^2:
\norm{\bar{\bm{w}}}\le w_0\}=\mathcal{K}^{*}$,
i.e.\ $\tilde{\bm{\lambda}},\tilde{\bm{\sigma}}\in\mathcal{K}$, while
preserving the primal--dual pairing
$\tilde{\bm{\lambda}}^\top\tilde{\bm{\sigma}}
=\bm{\lambda}^\top(\bm{\sigma}+\Gamma_\mu(\bm{\sigma}))$.
For $\tilde{\bm{\lambda}},\tilde{\bm{\sigma}}\in\mathcal{K}$, the scalar
complementarity condition can equivalently be expressed through the Jordan
product~\cite{AlizadehGoldfarb2003,Acary2026},
$\tilde{\bm{\lambda}}^\top\tilde{\bm{\sigma}}=0
\iff
\tilde{\bm{\lambda}}\circ\tilde{\bm{\sigma}}=\bm{0}$,
where $\circ$ denotes the Jordan product on $\mathcal{K}$
(Appendix~\ref{app:jordan}). The de~Saxc\'e condition~\eqref{eq:soccp} thus
becomes the second-order cone complementarity problem (SOCCP)
\begin{equation}\label{eq:jordansys}
\underbrace{\bm{\sigma}=A\bm{\lambda}+\bm{b}}_{\text{affine map}},
\qquad
\underbrace{\tilde{\bm{\lambda}},\tilde{\bm{\sigma}}\in\mathcal{K}}
_{\text{primal/dual feasibility}},
\qquad
\underbrace{\tilde{\bm{\lambda}}\circ\tilde{\bm{\sigma}}=\bm{0}}
_{\text{complementarity}},
\end{equation}
whose three solution branches are the separating, clamping, and sliding modes
of~\eqref{eq:modes}, and which can be solved by Jordan-based interior-point
and path-following methods~\cite{Acary2026}.

\subsection{Contact impulse sensitivity}\label{sec:bg_exact}
The contact impulse $\bm{\lambda}$ is implicitly determined by the Delassus
pair $(A,\bm{b})$ and the contact conditions in \eqref{eq:modes}.
Differentiating these conditions together with
$\bm{\sigma}=A\bm{\lambda}+\bm{b}$, while holding the contact mode fixed,
gives a separate sensitivity for each mode~\cite{Werling2021,LeLidec2025}.
For given variations $dA$ and $d\bm{b}$, the corresponding
$d\bm{\lambda}$ is obtained as follows.

\emph{Separating.} Within the separating mode, the contact impulse remains
zero, and therefore
\begin{equation}\label{eq:limit_brk}
d\bm{\lambda} = \bm{0}.
\end{equation}

\emph{Clamping.} The clamping condition is
$\bm{\sigma}=A\bm{\lambda}+\bm{b}=\bm{0}$. Differentiating this condition
gives
\begin{equation}\label{eq:limit_stk}
A\,d\bm{\lambda}
= -\bigl((dA)\bm{\lambda}+d\bm{b}\bigr),
\end{equation}
and hence
$d\bm{\lambda}
=-A^{-1}\bigl((dA)\bm{\lambda}+d\bm{b}\bigr)$.

\emph{Sliding.} In the sliding mode, the Signorini condition gives
$\sigN=0$, while the maximum-dissipation principle imposes
$\lamT=-\mu\lamN\,\hat{\bm{\sigma}}_T$. Following~\cite[App.~A]{LeLidec2025},
differentiating $\bm{\sigma}=A\bm{\lambda}+\bm{b}$ gives
$d\bm{\sigma}=A\,d\bm{\lambda}+(dA)\bm{\lambda}+d\bm{b}$, so the normal
condition gives
$A_N\,d\bm{\lambda}=-((dA)\bm{\lambda}+d\bm{b})_N$.
Differentiating the MDP relation through $\hat{\bm{\sigma}}_T$ gives
$K\,d\bm{\lambda}=-(1/\alpha)P\,d\sigT$, where
$K:=[\,\mu\hat{\bm{\sigma}}_T\ \ I_2\,]$,
$P:=I_2-\hat{\bm{\sigma}}_T\hat{\bm{\sigma}}_T^\top$, and
$\alpha:=\norm{\sigT}/(\mu\lamN)$. Eliminating $d\sigT$ using
$d\sigT=A_T\,d\bm{\lambda}+((dA)\bm{\lambda}+d\bm{b})_T$ gives
\begin{equation}\label{eq:sld}
\begin{aligned}
\sigN=0:\quad
&A_N\,d\bm{\lambda}
= -\bigl((dA)\bm{\lambda}+d\bm{b}\bigr)_N,\\[2pt]
\text{MDP}:\quad
&\left(\tfrac{1}{\alpha}P A_T+K\right)d\bm{\lambda}
= -\tfrac{1}{\alpha}P
\bigl((dA)\bm{\lambda}+d\bm{b}\bigr)_T .
\end{aligned}
\end{equation}
Here, $A_N$ and $A_T$ denote the normal and tangential rows of $A$.
The two conditions form a $3\times3$ linear system for
$d\bm{\lambda}\in\R^3$.

For multiple contacts, the mode-specific conditions are assembled into a
single linear system. Separating contacts satisfy
$d\bm{\lambda}=\bm{0}$ and are omitted from the reduced system, while the
clamping and sliding equations are stacked for the remaining contacts.
The off-diagonal Delassus blocks $A^{ij}$ couple the contact sensitivities,
and a single linear solve yields the remaining contact-impulse
variations~\cite{Werling2021,Kim2023,LeLidec2025}. Because differentiation is 
performed with the contact-mode assignment fixed,
the sensitivity is piecewise defined over $(A,\bm{b})$, with a different
expression for each mode assignment.

\begin{figure}[t]
\centering
\includegraphics[width=\columnwidth]{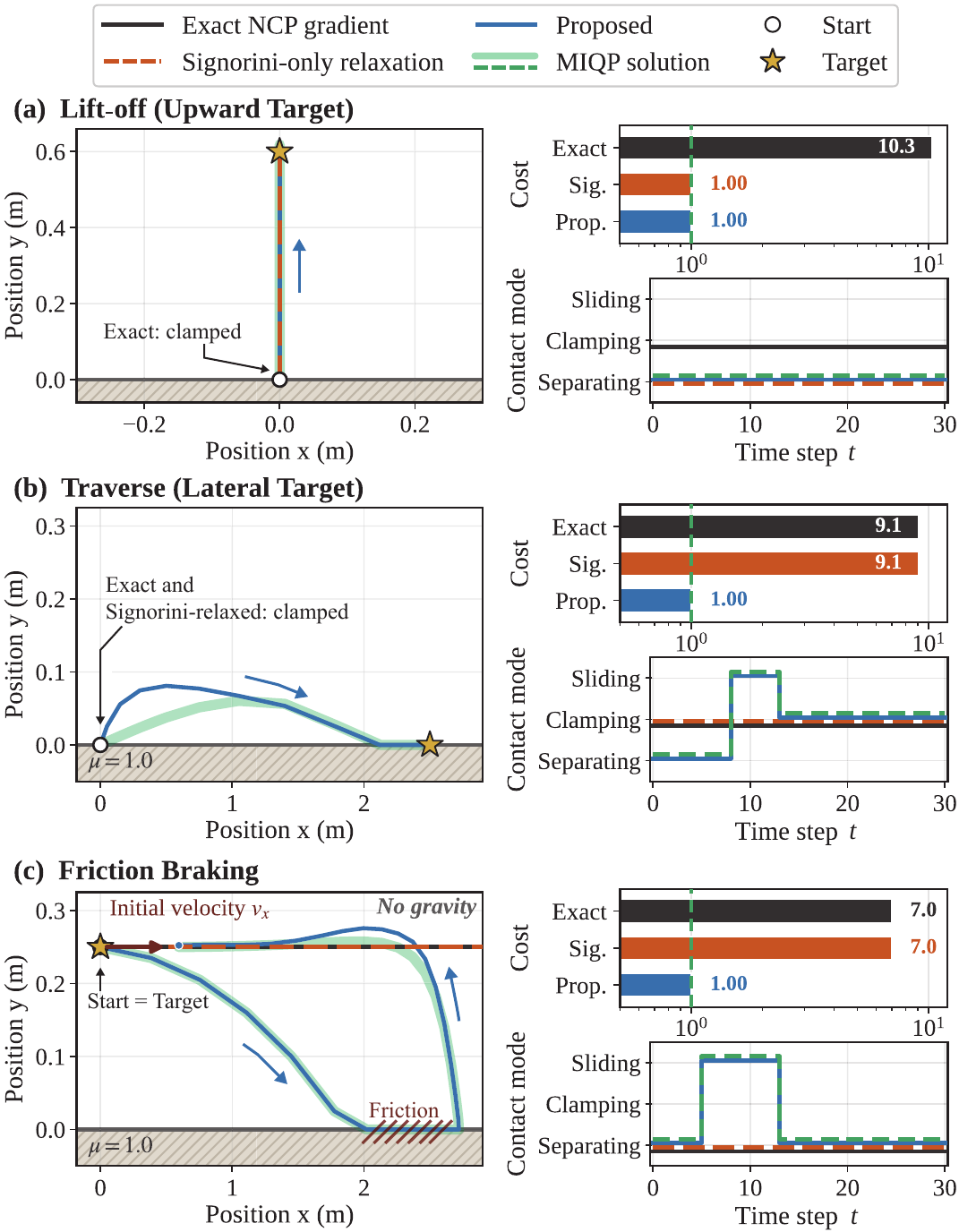}
\caption{Converged iLQR solutions for three CITO tasks over a 30-step horizon with hard-contact forward dynamics. The solutions differ only in the backward gradient: mode-dependent exact, Signorini-relaxed, or proposed full-cone relaxation. (a) Lift-off: the mass starts clamped on the ground with the target above its initial position. (b) Traverse: the mass starts clamped with a lateral target. (c) Friction braking: without gravity, the mass hovers above the ground at the target position with an initial lateral velocity and returns after moving away. All rollouts start from zero control. An MIQP over the contact-mode sequence, solved to zero optimality gap, provides the reference trajectory (dashed) and cost normalization. Right: final cost normalized by the MIQP cost and contact-mode sequence over the horizon; the green marker denotes the MIQP solution.}
\label{fig:pointmass}
\end{figure}

\section{Problem Setting}\label{sec:problem}\label{sec:bg_reqs}

In this section, we examine what contact sensitivities must capture to guide first-order optimization near transitions between contact modes. We use point-mass examples to study how different relaxations of the frictional contact condition affect the available optimization directions.

\subsection{Requirements on the backward contact gradient}
The exact contact sensitivity can be uninformative for first-order updates that require a contact-mode transition~\cite{Werling2021,LeLidec2024rs}; the optimizer can therefore remain confined to the current contact mode even when another mode would reduce the objective. The sensitivity is obtained by differentiating the contact constraints for the mode selected by the forward solve; it thus describes only perturbations that preserve that mode. For a loss $\ell$, the contribution to its derivative through contact-point motion can be written as $(\partial\ell/\partial\bm{\sigma})(d\bm{\sigma}/d\eta)$, where $\bm{\sigma}$ denotes the contact-point velocity. The first factor measures the sensitivity of the loss to contact-point motion, whereas $d\bm{\sigma}/d\eta$ is determined by the contact sensitivity. The limitation arises from the latter term. At a clamping contact, the mode constraint imposes $\bm{\sigma}=\bm{0}$, and differentiation within the fixed mode gives $d\bm{\sigma}/d\eta=\bm{0}$. Hence, the loss derivative transmitted through contact-point motion vanishes even when such motion would reduce the objective. Analogous restrictions arise in the other contact modes. At a separating contact, $\bm{\lambda}=\bm{0}$ gives $d\bm{\lambda}/d\eta=\bm{0}$; at a sliding contact, $\sigma_N=0$ and the differentiated MDP constrain the sensitivity to the sliding branch.

Fig.~\ref{fig:pointmass} considers three CITO tasks for a planar point mass actuated by horizontal and vertical forces. Each task uses \(m=1\,\mathrm{kg}\), a \(0.05\,\mathrm{s}\) time step, a horizon of \(H=30\), and friction coefficient \(\mu=1\). The objective is
\vspace{-0.2mm}
\begin{equation*}
\begin{aligned}
J={}&\sum_{t=0}^{H-1}\bigl[
\|\mathbf p_t-\mathbf p_d\|^2
+w_v\|\mathbf v_t\|^2
+w_u\|\mathbf u_t\|^2\bigr]\\
&+w_f\bigl[
\|\mathbf p_H-\mathbf p_d\|^2
+w_v\|\mathbf v_H\|^2\bigr],
\end{aligned}
\end{equation*}
where \(\mathbf p_t\), \(\mathbf v_t\), and \(\mathbf u_t\) are the position, velocity, and applied force at step \(t\), and \(\mathbf p_d\) is the target position. The weights are \(w_v=10^{-3}\), \(w_u=10^{-4}\), and \(w_f=20\). Within each task, all three methods use the same objective, force limit (20 N in (a) and (b); 6 N in (c)), zero-control initialization, iLQR solver, and hard-contact forward dynamics. Only the backward contact sensitivity differs: mode-dependent exact, Signorini-relaxed (normal-only), or proposed full-cone relaxation. Under zero-control initialization, cases (a) and (b) remain clamped under gravity, whereas case (c) starts above the ground in the separating mode without gravity and with an initial lateral velocity. For comparison, we solve a mixed-integer quadratic program (MIQP) that optimizes the contact-mode sequence to zero optimality gap. Its trajectory is shown in green, and the reported costs are normalized by the MIQP solution cost.

Fig.~\ref{fig:pointmass} shows that the proposed gradient reaches the target in all three tasks. The Signorini-relaxed gradient succeeds in (a) but not in (b) or (c), whereas the exact gradient does not reach the target in any task. In (a), reaching the target above the initial position requires a transition from clamping to separation. The exact sensitivity gives zero normal contact-velocity variation, so the contact remains clamped. Relaxing the normal contact condition is sufficient to reach the target. In (b), reaching the lateral target requires tangential contact-velocity variation. For the clamped initialization, both the exact and Signorini-relaxed sensitivities give zero tangential variation, yielding higher-cost trajectories that remain clamped in place. The proposed gradient yields a lower-cost trajectory following the MIQP strategy: the mass lifts off and travels laterally in the separating mode, avoiding frictional resistance along the ground, before sliding and reclamping at the target.

In (c), the target coincides with the initial position, but the initial lateral velocity carries the mass past it. The bounded actuation force cannot return the mass within the horizon; descending into ground contact introduces a frictional force that can further decelerate the mass. The exact and Signorini-relaxed sensitivities do not provide the coupled normal and tangential variation needed for this motion. Both optimizations saturate the force input without returning to the target. The proposed gradient yields a trajectory that descends, brakes through sliding friction, and returns to the target. Across these tasks, covering all normal and tangential contact-mode transitions
requires relaxing the full frictional contact condition.

\section{Analytic Contact Sensitivity via Relaxed de~Saxc\'e Conic Complementarity}

\begin{figure*}[t]
\centering
\includegraphics[width=\textwidth]{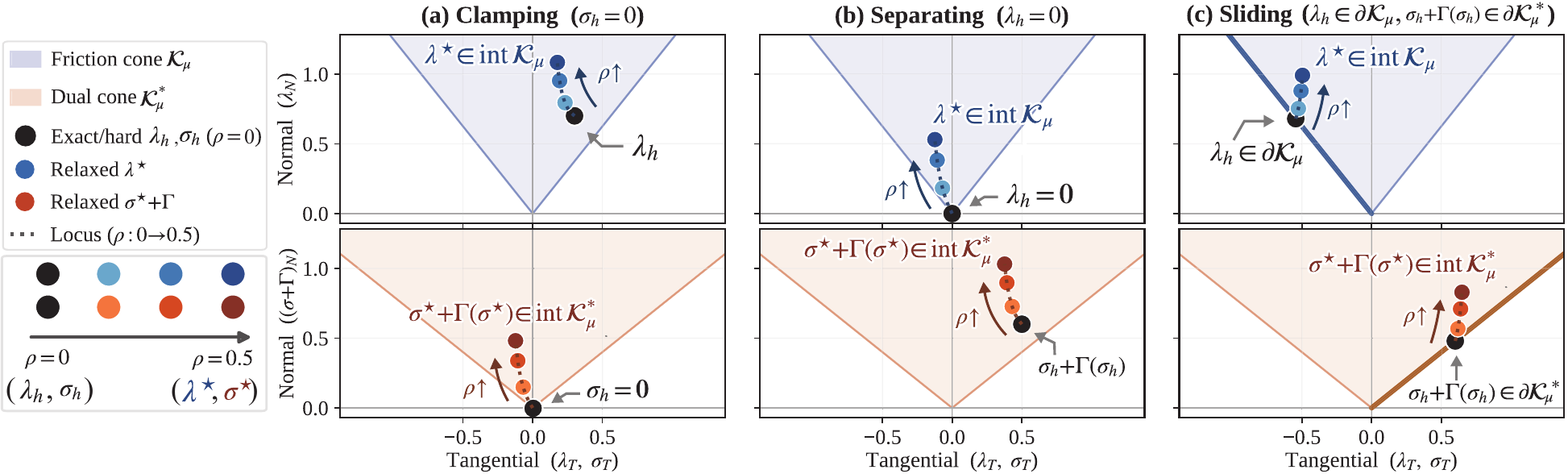}
\caption{Central-path relaxation of the exact contact solutions in the clamping, separating, and sliding regimes for a two-dimensional contact with $\mu=0.8$. The top row shows the contact impulse $\bm{\lambda}\in\Kmu$, and the bottom row shows the de~Saxc\'e-corrected contact velocity $\bm{\sigma}+\Gamma_\mu(\bm{\sigma})\in\Kmus$. At $\rho=0$, the clamping regime has $\bm{\sigma}_h=\bm{0}$, the separating regime has $\bm{\lambda}_h=\bm{0}$, and, in the sliding regime, both $\bm{\lambda}_h$ and $\bm{\sigma}_h+\Gamma_\mu(\bm{\sigma}_h)$ lie on their respective cone boundaries. For $\rho>0$, the corresponding relaxed variables lie in $\operatorname{int}\Kmu$ and $\operatorname{int}\Kmus$. Black markers denote the exact solution. Blue and orange markers denote the relaxed impulse $\bm{\lambda}^\star$ and corrected contact velocity $\bm{\sigma}^\star+\Gamma_\mu(\bm{\sigma}^\star)$, respectively, for $\rho\in(0,0.5]$. Darker markers correspond to larger $\rho$, and dotted curves trace the relaxed-solution loci as $\rho$ increases.}
\label{fig:regimes}
\end{figure*}

\label{sec:method}

The proposed contact sensitivity is obtained by relaxing the de~Saxc\'e-corrected conic complementarity along its central path at $\rho>0$ and analytically differentiating the relaxed relation. The relaxation couples the normal and tangential sensitivities without fixing the clamping, sliding, or separating mode. A Newton iteration initialized from the hard-contact solution computes a relaxed evaluation pair $(\bm{\lambda}^\star,\bm{\sigma}^\star)$ satisfying both the relaxed complementarity and the Delassus relation. A normal free-velocity offset $\delta$ additionally adjusts the clamping--separating sensitivity. Both modifications are restricted to the backward sensitivity evaluation.

\begin{figure*}[t]
\centering
\includegraphics[width=\textwidth]{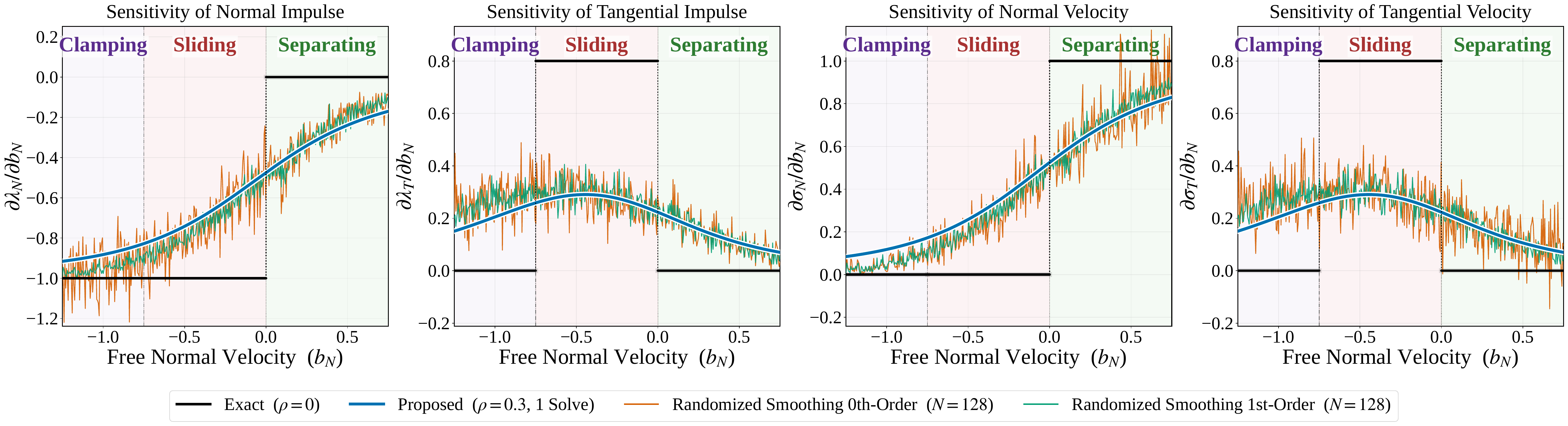}
\caption{Sensitivities of a single contact under the exact, randomized-smoothing, and proposed gradients: the impulse and contact-velocity components with respect to the free normal velocity $b_N$. The point-mass model has diagonal $A$ and nonzero free tangential velocity $b_T$. As $b_N$ increases, the hard-contact solution changes mode from clamping to sliding and then to separating. The exact sensitivity (black, $\rho=0$) jumps at the mode boundaries and contains zero components resulting from differentiation of the contact constraints with the mode held fixed. Randomized smoothing with perturbation standard deviation $0.6$ yields zeroth- and first-order estimates~\cite{Suh2022,LeLidec2024rs} using $N=128$ perturbed solves per point. The proposed sensitivity ($\rho=0.3$) is evaluated deterministically at a single Newton-solved central-path point for each $b_N$ (Section~\ref{sec:method_corrector}) and remains continuous across these mode transitions.}
\label{fig:sweep}
\end{figure*}

\subsection{Central-path relaxation and the unified contact sensitivity}\label{sec:method_sens}
The de~Saxc\'e-corrected conic complementarity $\tilde{\bm{\lambda}}\circ\tilde{\bm{\sigma}}=\bm{0}$ is relaxed along a central path, as in primal--dual interior-point methods for frictional contact formulated as an SOCCP~\cite{Acary2026}:
\begin{equation}\label{eq:relax}
\tilde{\bm{\lambda}}\circ\tilde{\bm{\sigma}}=\rho e,\qquad
\rho>0,\qquad e=[1,\bm{0}^\top]^\top,
\end{equation}
which places both $\tilde{\bm{\lambda}}$ and $\tilde{\bm{\sigma}}$ in $\mathrm{int}\,\mathcal{K}$. We differentiate this interior relation directly, deriving the contact sensitivity without distinguishing the separating, clamping, and sliding modes.

At fixed $\rho$, differentiating~\eqref{eq:relax} using the Jordan-product representation gives
\begin{equation}\label{eq:jordan_diff}
L(\tilde{\bm{\sigma}})\,d\tilde{\bm{\lambda}}
+L(\tilde{\bm{\lambda}})\,d\tilde{\bm{\sigma}}=\bm{0},
\end{equation} 
where $L(\cdot)$ is the Jordan-product arrow matrix (Appendix~\ref{app:jordan})~\cite{AlizadehGoldfarb2003}. For the scaled impulse,
\begin{equation*}
L(\tilde{\bm{\lambda}})
=\begin{bmatrix}
\mu\lamN & \bm{\lambda}_T^\top\\
\bm{\lambda}_T & \mu\lamN I_2
\end{bmatrix},
\end{equation*}
with $L(\tilde{\bm{\sigma}})$ defined analogously. From $\tilde{\bm{\lambda}}=[\,\mu\lamN,\bm{\lambda}_T^\top\,]^\top$ and $\tilde{\bm{\sigma}}=[\,\sigN/\mu+\norm{\sigT},\bm{\sigma}_T^\top\,]^\top$, differentiating these scalings gives $d\tilde{\bm{\lambda}}=C\,d\bm{\lambda}$ and $d\tilde{\bm{\sigma}}=D(\bm{\sigma})\,d\bm{\sigma}$, where
\begin{equation*}
C=\begin{bmatrix}\mu&\bm{0}^\top\\ \bm{0}&I_2\end{bmatrix},
\qquad
D(\bm{\sigma})=
\begin{bmatrix}
1/\mu & \bm{\sigma}_T^\top/\norm{\bm{\sigma}_T}\\
\bm{0} & I_2
\end{bmatrix}.
\end{equation*}
Substituting these differentials into~\eqref{eq:jordan_diff} gives
\begin{equation}\label{eq:compact}
\begin{split}
\Mlam\,d\bm{\lambda}+\Msig\,d\bm{\sigma}&=\bm{0},\\
\Mlam:=L(\tilde{\bm{\sigma}})\,C,\quad
&\Msig:=L(\tilde{\bm{\lambda}})\,D(\bm{\sigma}),
\end{split}
\end{equation}
where the explicit block forms of $\Mlam$ and $\Msig$ are given as follows. 

Using the definitions of $L(\cdot)$, $C$, and $D(\bm{\sigma})$,\footnote{For $\sigT=\bm{0}$, we set $\sigT/\norm{\sigT}=\bm{0}$. This convention applies at the initialization of Algorithm~\ref{alg:newton} from a clamping contact, where $\bm{\sigma}_h=\bm{0}$, and at any iterate with $\sigT=\bm{0}$.} direct multiplication of
$\Mlam=L(\tilde{\bm{\sigma}})C$ and
$\Msig=L(\tilde{\bm{\lambda}})D(\bm{\sigma})$ gives, 
\begin{equation*}
\Msig = \left[\begin{array}{c|c}
\lamN & \mu\lamN\dfrac{\sigT^\top}{\norm{\sigT}}+\lamT^\top \\[6pt]\hline\\[-6pt]
\dfrac{\lamT}{\mu} & \lamT\dfrac{\sigT^\top}{\norm{\sigT}}+\mu\lamN I_2
\end{array}\right],
\end{equation*}
and
\begin{equation*}
\Mlam = \left[\begin{array}{c|c}
\sigN+\mu\norm{\sigT} & \sigT^\top \\[4pt]\hline\\[-6pt]
\mu\sigT & \bigl(\sigN/\mu+\norm{\sigT}\bigr)I_2
\end{array}\right].
\end{equation*}

Combining with the differentiated Delassus relation $d\bm{\sigma}=A\,d\bm{\lambda}+(dA)\bm{\lambda}+d\bm{b}$ gives the unified sensitivity equation
\begin{equation}\label{eq:sens}
(\Mlam+\Msig A)\,d\bm{\lambda}
=-\Msig\bigl((dA)\bm{\lambda}+d\bm{b}\bigr).
\end{equation}
For a single contact, for example, the sensitivity with respect to a variable $\eta$ is
\begin{equation*}
\frac{\partial\bm{\lambda}}{\partial\eta}
=-(\Mlam+\Msig A)^{-1}\Msig
\left(\frac{\partial A}{\partial\eta}\bm{\lambda}
+\frac{\partial\bm{b}}{\partial\eta}\right).
\end{equation*}
For $n_c$ contacts, the per-contact $3\times3$ blocks of $\Mlam$ and $\Msig$ assemble into the corresponding $3n_c$-dimensional system, which is solved jointly for the contact-impulse sensitivity. Taking $\eta$ as a component of $\bm{x}$ or $\bm{u}$ provides the contact sensitivities used to compute $f_{\bm{x}}$ and $f_{\bm{u}}$.

\subsection{Evaluating the sensitivity at $\rho>0$}
\label{sec:method_corrector}
We evaluate the sensitivity in~\eqref{eq:sens} at a relaxed pair $(\bm{\lambda}^\star,\bm{\sigma}^\star)$ satisfying the Delassus relation $\bm{\sigma}=A\bm{\lambda}+\bm{b}$ and the relaxed complementarity $\tilde{\bm{\lambda}}\circ\tilde{\bm{\sigma}}=\rho e$ for $\rho>0$, rather than at the hard-contact pair. This change in the evaluation point also changes the constraints imposed on the linearized contact response. At $\rho=0$, the hard-contact pair satisfies rigid complementarity and lies on a mode-dependent boundary of the cones. Differentiation with the contact mode fixed enforces the conditions in~\eqref{eq:modes}; for example, $d\bm{\sigma}=\bm{0}$ in clamping and $d\bm{\lambda}=\bm{0}$ in separating. The exact sensitivity therefore provides no first-order direction toward a change in the current contact mode. For $\rho>0$, a relaxed pair on the physical central-path branch satisfies $(\tilde{\bm{\lambda}}^\star,\tilde{\bm{\sigma}}^\star) \in\mathrm{int}\,\mathcal{K}\times\mathrm{int}\,\mathcal{K}$. Its linearization is governed by the unified system~\eqref{eq:sens} rather than the hard-contact mode conditions, so the corresponding sensitivity components can be nonzero. 

Fig.~\ref{fig:regimes} illustrates the evaluation-point geometry for clamping, separating, and sliding contacts. At the hard-contact root, $\tilde{\bm{\sigma}}_h=\bm{0}$ in clamping and $\tilde{\bm{\lambda}}_h=\bm{0}$ in separating, whereas both variables are nonzero and lie on their respective cone boundaries in sliding. As $\rho$ increases from zero, the relaxed points depart from their hard-contact roots and move into their respective cone interiors. Fig.~\ref{fig:sweep} shows the sensitivities for a single-contact model with diagonal $A$ and nonzero $b_T$ as $b_N$ is swept across the three contact modes. For the exact sensitivity, the tangential impulse and contact-velocity derivatives are zero in clamping and separating, and the normal contact-velocity derivative is zero in clamping and sliding. These derivatives change discontinuously at the contact-mode boundaries. For $\rho>0$, the proposed curves vary continuously across the plotted contact-mode transitions, and components constrained to zero by the exact contact mode become nonzero; in particular, the full-cone coupling produces nonzero tangential sensitivities for a normal perturbation, such as $db_N$ in this example. The empirical means obtained from repeated zeroth- and first-order randomized-smoothing estimates show the same qualitative features in this sweep~\cite{Suh2022,LeLidec2024rs}, while the proposed sensitivity does not require Monte Carlo sampling.

The relaxed evaluation point is computed by a Newton iteration initialized at the hard-contact solution $(\bm{\lambda}_h,\bm{\sigma}_h)$ returned by the forward pass, while the associated Delassus pair $(A,\bm b)$ is held fixed. To cast the relaxed complementarity together with the Delassus relation as a root-finding problem in the impulse alone, we substitute $\bm{\sigma}=A\bm{\lambda}+\bm b$ into~\eqref{eq:relax} and define the residual \begin{equation}\label{eq:Fdef}
\Frho(\bm{\lambda}):=\tilde{\bm{\lambda}}(\bm{\lambda})\circ\tilde{\bm{\sigma}}\bigl(A\bm{\lambda}+\bm b\bigr)-\rho\bm e.
\end{equation} 
Its Jacobian is the system matrix in~\eqref{eq:sens},
\begin{equation*}
\frac{\partial\Frho}{\partial\bm{\lambda}}=L(\tilde{\bm{\sigma}})\,C+L(\tilde{\bm{\lambda}})\,D(\bm{\sigma})\,A=\Mlam+\Msig A.
\end{equation*}
To solve $\Frho(\bm{\lambda})=\bm 0$, we use the following Newton update with a regularization term $\beta\norm{\Frho^{(k)}}I$ at iteration $k$:
\begin{equation}\label{eq:newton}
\begin{aligned}
\bigl(\Mlam^{(k)}+\Msig^{(k)}A+\beta\norm{\Frho^{(k)}}I\bigr)\Delta\bm{\lambda}^{(k)}&=-\Frho^{(k)},\\
\bm{\lambda}^{(k+1)}&=\bm{\lambda}^{(k)}+\Delta\bm{\lambda}^{(k)},\\
\bm{\sigma}^{(k+1)}&=A\bm{\lambda}^{(k+1)}+\bm b,
\end{aligned}
\end{equation}
where $\Frho^{(k)}=\Frho(\bm{\lambda}^{(k)})$. The last update keeps the Delassus relation satisfied at every iterate. Upon convergence, $(\bm{\lambda}^\star,\bm{\sigma}^\star)$ satisfies both relations and is used to evaluate $\Mlam$ and $\Msig$ for the contact sensitivity.

The term $\beta\norm{\Frho^{(k)}}I$ provides a residual-scaled diagonal regularization of the Newton system, as in regularized Newton methods for nonlinear equations~\cite{xiao2018regularized}; here, $\beta\geq0$ is the regularization parameter. This term regularizes the linear solve during the early Newton iterations, particularly when the hard-contact initialization yields a poorly conditioned Newton matrix. For example, for a clamping contact close to the friction-cone boundary, $\mu\lambda_N-\norm{\bm{\lambda}_T}$ becomes small and $L(\tilde{\bm{\lambda}})$ approaches singularity.\footnote{The eigenvalues of $L(\tilde{\bm{\lambda}})$ include $\mu\lambda_N-\norm{\bm{\lambda}_T}$, which approaches zero at the clamping--sliding boundary.} As the Newton iteration converges and the residual decreases, the regularization weakens and vanishes at the relaxed root, recovering the unregularized Newton system. Algorithm~\ref{alg:newton} summarizes the procedure.

\begin{algorithm}[t]
\caption{Regularized Newton iteration for the backward evaluation point}
\label{alg:newton}
\begin{algorithmic}[1]
\Require Hard-contact impulse $\bm{\lambda}_h$; Delassus pair $(A,\bm b)$;
$\rho>0$; regularization parameter $\beta\geq0$; tolerance $\varepsilon$;
maximum iterations $k_{\max}$
\State $\bm{\lambda}\leftarrow\bm{\lambda}_h$;\quad
$\bm{\sigma}\leftarrow A\bm{\lambda}+\bm b$
\For{$k=0,\dots,k_{\max}-1$}
  \State Form $\tilde{\bm{\lambda}},\tilde{\bm{\sigma}}$ and evaluate
  $\Frho$ from~\eqref{eq:Fdef}
  \If{$\norm{\Frho}<\varepsilon$}
    \State \textbf{break}
  \EndIf
  \State Assemble $\Mlam,\Msig$ from~\eqref{eq:compact} and solve
  $(\Mlam+\Msig A+\beta\norm{\Frho}I)\Delta\bm{\lambda}=-\Frho$ as in~\eqref{eq:newton}
  \State $\bm{\lambda}\leftarrow\bm{\lambda}+\Delta\bm{\lambda}$;\quad
  $\bm{\sigma}\leftarrow A\bm{\lambda}+\bm b$
\EndFor
\State \Return $(\bm{\lambda}^\star,\bm{\sigma}^\star)
\leftarrow(\bm{\lambda},\bm{\sigma})$
\end{algorithmic}
\end{algorithm}

\begin{figure}[t]
\centering
\includegraphics[width=\columnwidth]{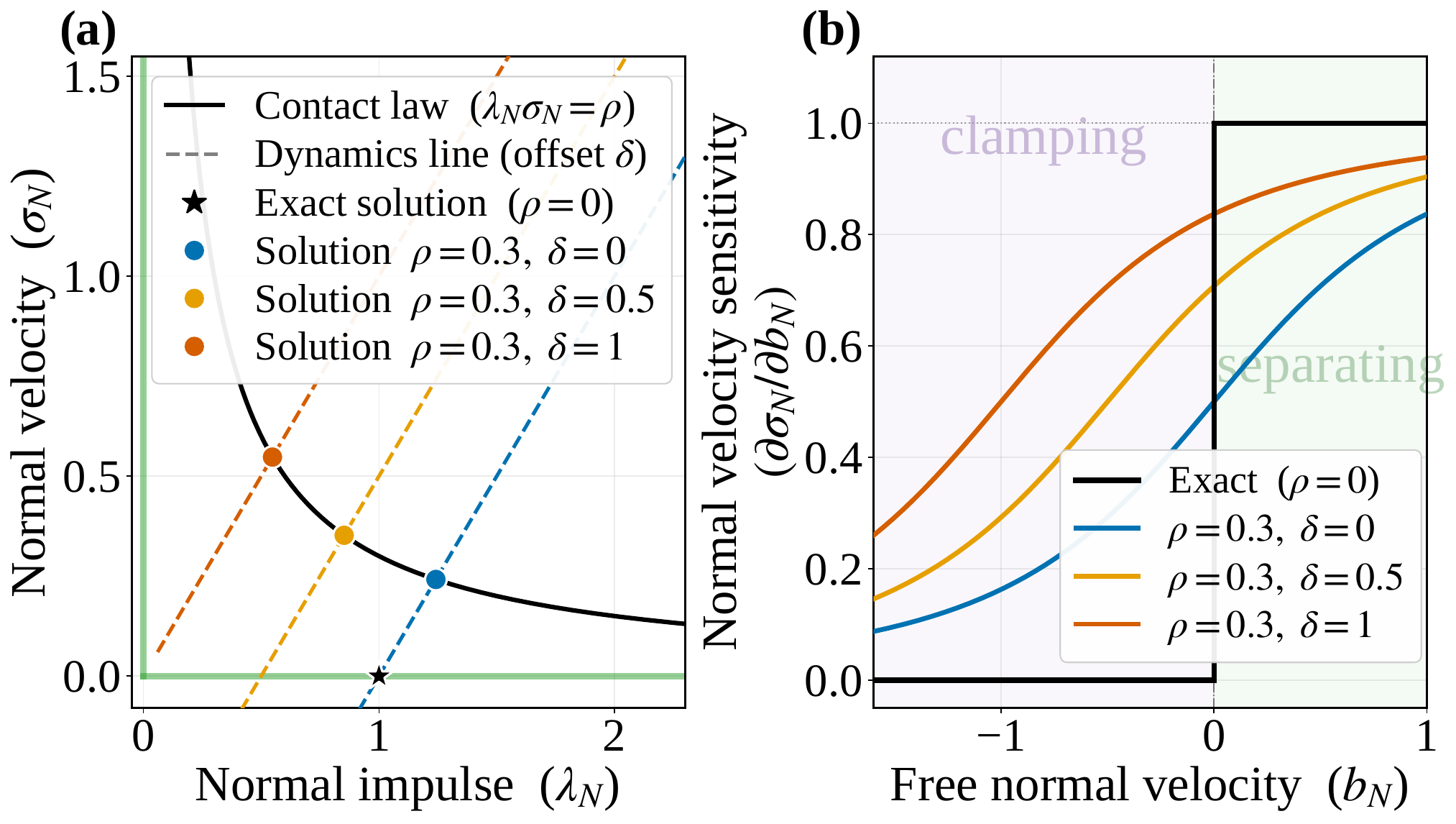}
\caption{Effect of the normal free-velocity offset $\delta$ in the proposed relaxation. (a)~Increasing $\delta$ translates the dynamics relation (dashed). Its intersection with the relaxed contact law $\lamN\sigN=\rho$ moves along the curve toward lower $\lamN$ and higher $\sigN$. (b)~Normal sensitivity $\partial\sigN/\partial b_N$ as a function of the unshifted $b_N$. Increasing $\delta$ shifts the sensitivity curve to more negative $b_N$.}
\label{fig:delta}
\end{figure}

\begin{figure*}[t]
\centering
\includegraphics[width=\textwidth]{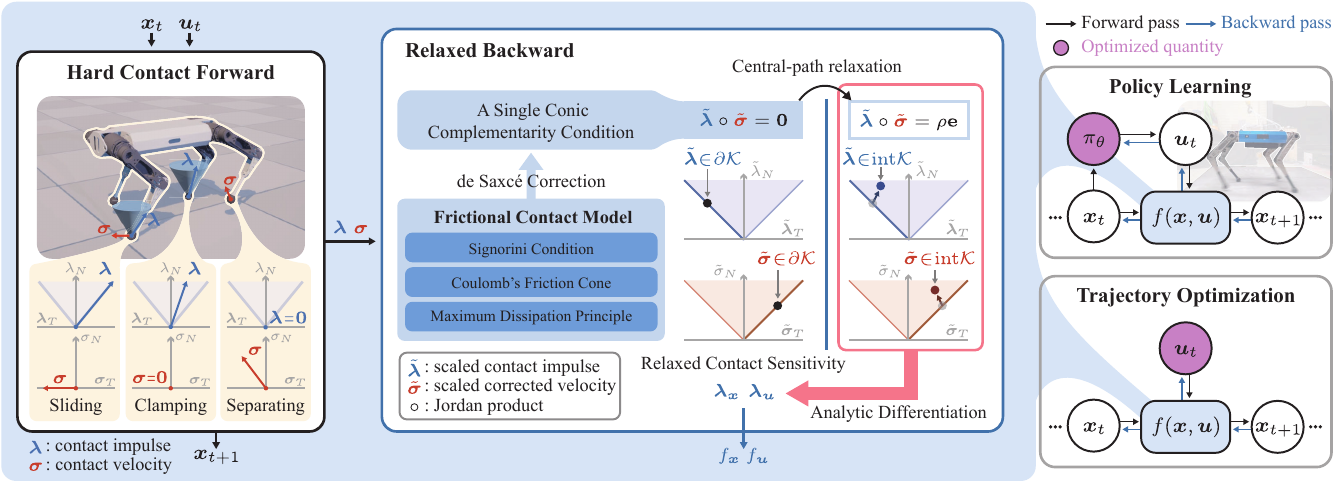}
\caption{Overview of the proposed differentiable contact framework. Given $(x_t,u_t)$, the hard-contact forward solve determines the contact impulse $\lambda$, contact velocity $\sigma$, and next state $x_{t+1}$ under the realized sliding, clamping, or separating mode. For the backward pass, the Signorini condition, Coulomb friction cone, and maximum-dissipation principle are expressed as a single conic complementarity condition using the de Saxc\'e correction. The hard-contact solution at $\rho=0$ is relaxed along the central path to $\rho>0$, and analytic differentiation yields the relaxed contact sensitivity and transition Jacobians $(f_x,f_u)$. These Jacobians are used for BPTT policy learning and DDP-based trajectory optimization.}
\label{fig:pipeline}
\end{figure*}

\subsection{Normal free-velocity offset}\label{sec:delta}
To improve numerical conditioning, we additionally apply a nonnegative offset $\delta$ to the normal component of the backward free-contact term before the
Newton iteration,
\begin{equation*}
b_N \leftarrow b_N+\delta.
\end{equation*}
With $\delta$ held fixed during differentiation, the offset changes the Newton residual and the relaxed evaluation pair $(\bm{\lambda}^\star,\bm{\sigma}^\star)$ without changing the form of the sensitivity system~\eqref{eq:sens}. For a given impulse, the offset adds a positive diagonal contribution to the Newton matrix, which can improve the conditioning of the linear solve. We observe a lower rate of convergence to spurious roots with the offset enabled (Section~\ref{sec:bptt_newton}).

Fig.~\ref{fig:delta} illustrates how increasing $\delta$ shifts the normal sensitivity curve toward more negative values of the unshifted $b_N$. At the same $b_N$, this shift provides greater normal sensitivity toward contact release for states whose hard-contact forward solution remains clamped. This sensitivity can guide trajectory optimization toward solutions involving liftoff.

\subsection{Recovery of the fixed-mode sensitivity as $\rho\to0$}\label{sec:method_limits}
In the limit $\rho\to0^+$, with $\delta=0$, the central-path relaxation vanishes
and the proposed sensitivity recovers the exact fixed-mode sensitivity derived
in Section~\ref{sec:bg_exact}. For a separating contact,
$\tilde{\bm{\lambda}}\to\bm{0}$ gives $\Msig\to0$ with nonsingular $\Mlam$,
so the limiting sensitivity gives $d\bm{\lambda}=\bm{0}$; for a clamping contact,
$\tilde{\bm{\sigma}}\to\bm{0}$ gives $\Mlam\to0$ with nonsingular $\Msig$,
so it gives $d\bm{\sigma}=\bm{0}$, equivalently
$A\,d\bm{\lambda}=-((dA)\bm{\lambda}+d\bm{b})$. These recover
\eqref{eq:limit_brk} and~\eqref{eq:limit_stk}, respectively. For a sliding
contact, substituting the hard solution into $\Mlam$ and $\Msig$ gives
\begin{equation*}
\Mlam\to\norm{\sigT}\begin{bmatrix}\mu&\unit^\top\\ \mu\unit&I_2\end{bmatrix},
\quad
\Msig\to\mu\lamN\begin{bmatrix}1/\mu&\bm{0}^\top\\ -\unit/\mu&P\end{bmatrix},
\end{equation*}
where $P:=I_2-\unit\unit^\top$. In
$\Mlam\,d\bm{\lambda}+\Msig\,d\bm{\sigma}=\bm{0}$, combining the first row
with the component of the last two rows along $\unit$ gives $d\sigN=0$.
With $d\sigN=0$, the last two rows reduce to
$K\,d\bm{\lambda}=-(1/\alpha)P\,d\sigT$, where
$K:=[\,\mu\unit\ \ I_2\,]$ and $\alpha:=\norm{\sigT}/(\mu\lamN)$.
These conditions coincide with the differentiated Signorini and MDP conditions
in~\eqref{eq:sld}, recovering the exact sliding sensitivity.

\section{Implementation}\label{sec:impl}\label{sec:apg_impl}\label{sec:method_compose}

Fig.~\ref{fig:pipeline} summarizes the forward--backward computation.
The forward pass determines the realized transition, while the
proposed relaxed contact sensitivity is used to assemble the transition
Jacobians for BPTT and DDP.

\subsection{Transition-Jacobian assembly}
The proposed gradient replaces the contact sensitivity used to form the
transition Jacobians. The semi-implicit step~\eqref{eq:vplus} defines the
discrete transition $\bm{x}_{t+1}=f(\bm{x}_t,\bm{u}_t,\bm{\lambda})$.
Differentiating the transition gives, for example,
\begin{equation}\label{eq:fxfu}
f_x =
\frac{\partial f}{\partial \bm{x}_t}
+
\frac{\partial f}{\partial \bm{\lambda}}\,
\frac{\partial \bm{\lambda}}{\partial \bm{x}_t},
\end{equation}
with an analogous expression for $f_u$. For the proposed backward pass,
$\partial \bm{\lambda}/\partial \bm{x}_t$ and $\partial  \bm{\lambda}/\partial \bm{u}_t$ are obtained from
\eqref{eq:sens} at the $\rho>0$ central-path evaluation point
$(\bm{\lambda}^\star,\bm{\sigma}^\star)$ obtained by
Algorithm~\ref{alg:newton}; the remaining factors, $\partial f/\partial\bm{x}_t$ at fixed impulse and $\partial f/\partial\bm{\lambda}$, are evaluated at the same point.

The other derivatives required to form $(f_x,f_u)$ are computed analytically with Pinocchio~\cite{Pinocchio}. The rigid-body derivatives are evaluated with the contact impulse held fixed, while the variations $dA$ and $d\bm{b}$ in \eqref{eq:sens} additionally require derivatives of the contact Jacobian, including the kinematic Hessian $\partial\Jc/\partial\bm{q}$. The derivative of the position-level correction included in $\bm{b}$ is also retained in $d\bm{b}$. Further details of the rigid-body derivative assembly are given in~\cite[App.~B]{Kim2023}.

\subsection{Optimal-control setup}
We evaluate policy learning and contact-implicit trajectory optimization on a
45-kg floating-base quadruped with a 37-dimensional state
($n_q=19$, $n_v=18$) and 12 joint-torque inputs. 
Contact is modeled at four point feet on flat ground under Coulomb friction, and the control step is 25~ms.
The transition Jacobians $(f_x,f_u)$ enter the BPTT recursion~\eqref{eq:bptt} for policy learning and the DDP/iLQR backward pass~\eqref{eq:ddp_Q} for trajectory optimization. The hard-contact forward solve is retained in both settings; the $\rho>0$ complementarity relaxation is applied only to the backward contact sensitivity.
   
\emph{Cost.} The regulating cost is
\begin{equation}\label{eq:cost}
\ell_{\mathrm{reg}}(\bm{x},\bm{u})
=
\|\bm{x}\ominus\bm{x}_{\mathrm{ref}}\|_{W_x}^{2}
+\|\bm{u}\|_{W_u}^{2},
\end{equation}
where $\bm{x}_{\mathrm{ref}}$ specifies the target base pose and velocity
together with the nominal joint configuration, and $W_x$ and $W_u$ are
diagonal weights. The floating-base orientation residual uses the $SO(3)$
logarithm, while the remaining state residuals are Euclidean
differences~\cite{Kim2023}. The state reference does not prescribe a contact
mode or contact sequence.

The policy-learning stage cost is
\begin{equation*}
\ell^{\mathrm{PL}}
=
\ell_{\mathrm{reg}}
+\ell_{\mathrm{foot}}
+\ell_{\mathrm{gait}}
+\ell_{\mathrm{smooth}}.
\end{equation*}
The foot term $\ell_{\mathrm{foot}}=\sum_i\|\bm{r}_i-\bar{\bm{r}}_i\|^2$ regulates the hip-relative horizontal foot position $\bm{r}_i$ about its nominal value $\bar{\bm{r}}_i$. For the gait-conditioned tasks, $\ell_{\mathrm{gait}}=\sum_i(\phi_i-\bar{\phi}_i)^2$ tracks a phase-dependent target $\bar{\phi}_i$ for the foot clearance $\phi_i$, with a maximum of $0.12$~m during swing and zero during stance; its weight is zero in the gait-free tasks. The smoothness term $\ell_{\mathrm{smooth}}=w_{\Delta a}\|\bm{a}_t-\bm{a}_{t-1}\|^2 +w_{\ddot q}\|\ddot{\bm{q}}_t\|^2$ penalizes changes in the action $\bm{a}_t$ and the joint acceleration $\ddot{\bm{q}}_t$, with weights $w_{\Delta a}$ and $w_{\ddot q}$. Trajectory optimization retains the cost weights of~\cite{Kim2023}, weighted toward position tracking. 

\begin{table}[t]
\caption{Running-cost weights for policy learning}
\label{tab:cost}
\centering
\footnotesize
\setlength{\tabcolsep}{2.5pt}
\begin{tabular*}{\columnwidth}{@{\extracolsep{\fill}}l r l r@{}}
\toprule
Term & Weight & Term & Weight \\
\midrule

\itshape Base & & \itshape Joints & \\
Position, $x,y$               & $0$
& Position                    & $5\times10^{-1}$ \\
Height                        & $4\times10^{1}$
& Hip-roll position           & $1$ \\
Orientation, roll--pitch      & $2\times10^{1}$
& Velocity                    & $5\times10^{-3}$ \\

\noalign{\vskip-.6\aboverulesep}
\cmidrule(l){3-4}
\noalign{\vskip-.6\belowrulesep}
Orientation, yaw              & $0$
& \itshape Feet               & \\
Linear velocity, $x,y$        & $4$
& Position, $x$               & $3$ \\
Linear velocity, $z$          & $2$
& Position, $y$               & $10^{1}$ \\
Angular velocity, roll--pitch & $2$
& Foot clearance$^\dagger$    & $10^{2}$ \\

\noalign{\vskip-.6\aboverulesep}
\cmidrule(l){3-4}
\noalign{\vskip-.6\belowrulesep}
Angular velocity, yaw         & $2$
& \itshape Regularization     & \\
                              &
& Control effort              & $5\times10^{-2}$ \\
                              &
& Action rate                 & $10^{-1}$ \\
                              &
& Joint acceleration          & $10^{-3}$ \\

\bottomrule
\end{tabular*}

\vspace{1pt}
\parbox{\columnwidth}{\scriptsize
$^\dagger$ Used only for gait-conditioned tasks.}
\end{table}

\subsection{Policy-learning implementation}
For policy learning, we consider velocity-command-tracking tasks in which the policy is conditioned on commands for planar linear velocity and yaw rate. The decision variable is the policy parameter $\theta$, which defines a feedback mapping from the current observation to an action.

The forward rollout uses RaiSim~\cite{Hwangbo2018} with a 1-ms simulation time step, yielding 25 simulation steps per control step at 40~Hz. Rollouts are differentiated over windows of $H=20$ control steps (0.5~s), as in short-horizon differentiable-simulation training~\cite{Xu2022,Song2024}. For the proposed backward pass, Algorithm~\ref{alg:newton} is initialized with the substep-accumulated impulse $\bm{\lambda}_h$ and with $\bm{\sigma}_h$ assembled at the simulated next state $\bm{x}_{t+1}$; its tangential components are the contact-point velocities, and its normal is the contact gap divided by the control step. The free-contact term of the backward Delassus pair is accordingly set to $\bm b:=\bm{\sigma}_h-A\bm{\lambda}_h$. After each policy update, the next training window continues from the terminal state of the preceding window; the simulator is reset only at episode termination. Episodes terminate after 4~s or when either the per-step cost or cumulative body-contact count exceeds its prescribed threshold. Adam~\cite{kingma2014adam} updates $\theta$ at a constant learning rate of $2\times10^{-3}$. The policy is deterministic.

\begin{figure*}[t]
\centering
\includegraphics[width=\textwidth]{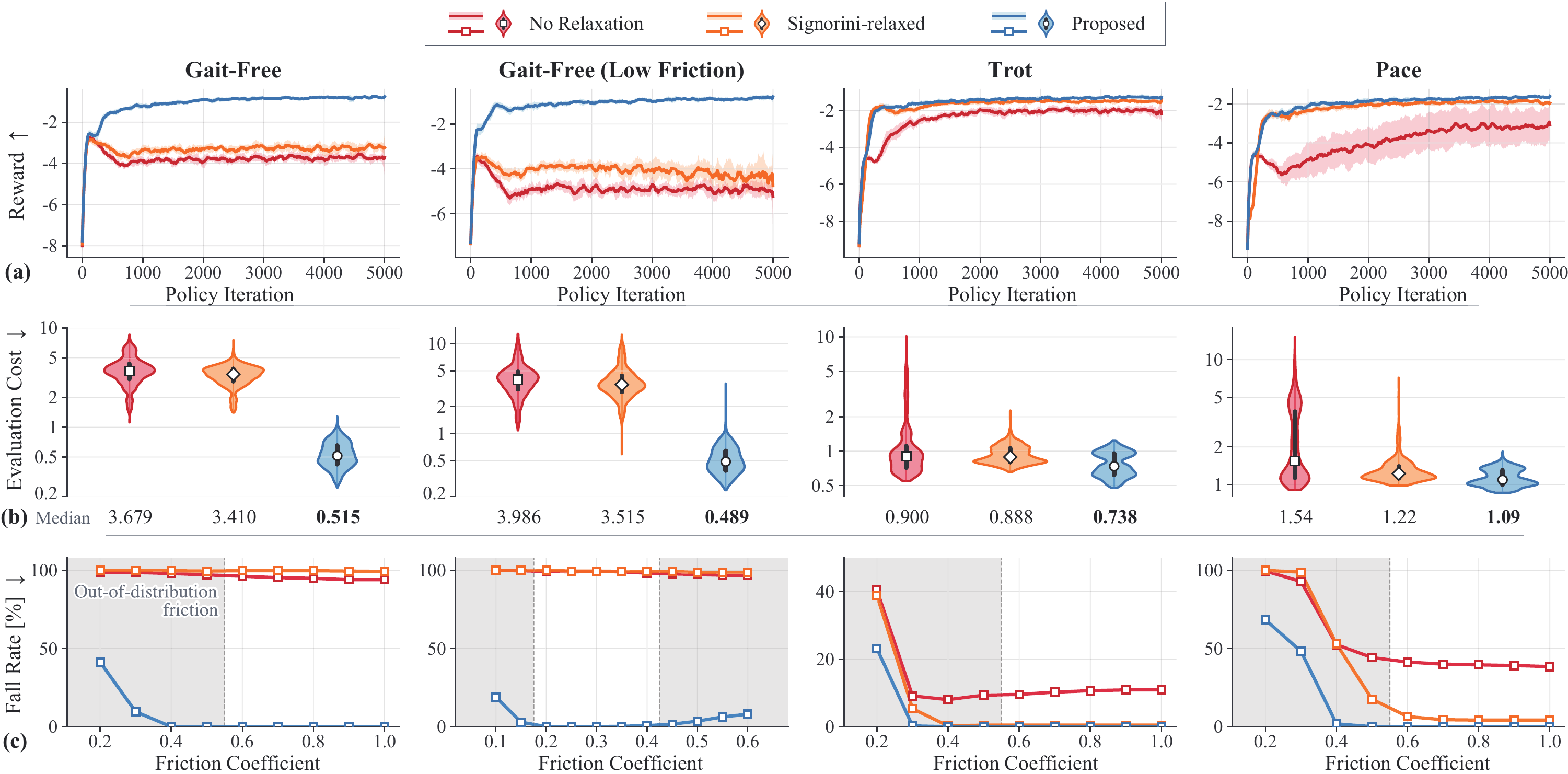}
\caption{Training with three backward contact gradients (no relaxation, Signorini-relaxed, and proposed) on four locomotion tasks (10 seeds). Training uses 16 environments, with friction coefficient $\mu\in[0.2,0.4]$ for low-friction gait-free locomotion and $\mu\in[0.6,1.0]$ otherwise. (a) Training reward. Curves show the mean and shaded bands denote $\pm1$ standard deviation across training curves smoothed separately for each seed. (b) Violin plots of evaluation cost. Each violin shows evaluation costs from 2,250 in-distribution rollouts across commands, friction coefficients, and seeds. Each cost is averaged over pre-fall steps. White markers and thick bars denote medians and interquartile ranges, respectively. Median costs are listed below the violins, with the lowest per task in bold. (c) Evaluation fall rate over an extended friction range; gray shading denotes out-of-distribution friction.}
\label{fig:apg_main}
\end{figure*}

The policy network uses a [256, 128] multilayer perceptron (MLP) with
LeakyReLU activations. Its 60-dimensional observation comprises the command,
gravity direction, joint positions and velocities, body twist, body-frame
foot positions, and previous action; gait-conditioned tasks add a
two-dimensional phase encoding. The policy observation is detached from the
computational graph before policy evaluation. The output $\bm{a}$ specifies
joint targets $\bm{q}^\star=\bm{q}_{\mathrm{nom}}+0.2\,\bm{a}$, which are
mapped to joint torques by PD gains $k_p=30$ and $k_d=1$, yielding a constant
action-to-torque Jacobian. Training is performed on a desktop with a 10-core
Intel Core i9-10900K CPU.

\subsection{Trajectory optimization implementation}\enspace 
In CITO, we consider base-pose reaching tasks. The reference specifies the desired base position and orientation while retaining the nominal joint configuration. Because no contact modes are preplanned, the optimized state--control trajectory determines the contact modes, contact impulses, and foothold placements required to reach the reference. For a given current state, the decision variable is the control sequence. Box-FDDP~\cite{mastalli2022feasibility} solves the resulting optimal-control problem over $H=20$ steps (0.5 s). For the nominal CITO evaluation, successive finite-horizon problems are solved in a receding-horizon fashion: the state is advanced one step along the optimized trajectory, which is then shifted to initialize the next solve.

\section{Policy Learning via Backpropagation Through Time}\label{sec:apg}

We evaluate how the backward contact gradient affects policy learning across the locomotion tasks. Hardware experiments assess policies trained with single-environment BPTT and deployed without fine-tuning. The PPO comparison evaluates sample efficiency and performance under a shared task objective and simulation setting. The Newton iteration is evaluated in terms of convergence and cone feasibility.

\begin{figure}[t]
\centering
\includegraphics[width=\columnwidth]{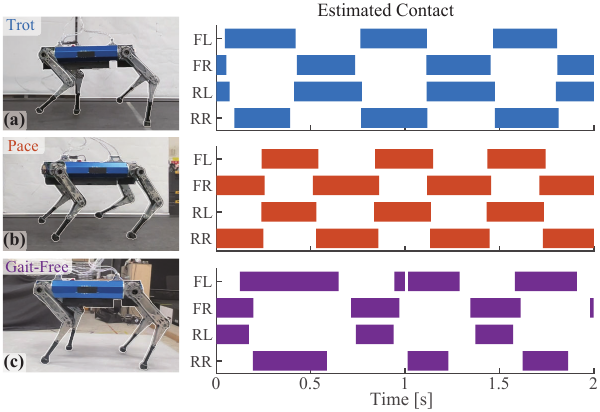}
\caption{Hardware deployment of BPTT-trained locomotion policies. Snapshots show the hardware experiments, and the plots show estimated foot-contact patterns. (a) Trot, (b) pace, and (c) gait-free locomotion. Training uses 100,000 control steps in 108~s for the trot and pace policies, and 160,000 steps in 172~s for the gait-free policy.}
\label{fig:bptt_exp_gait}
\end{figure}

\begin{figure*}[t]
\centering
\includegraphics[width=\textwidth]{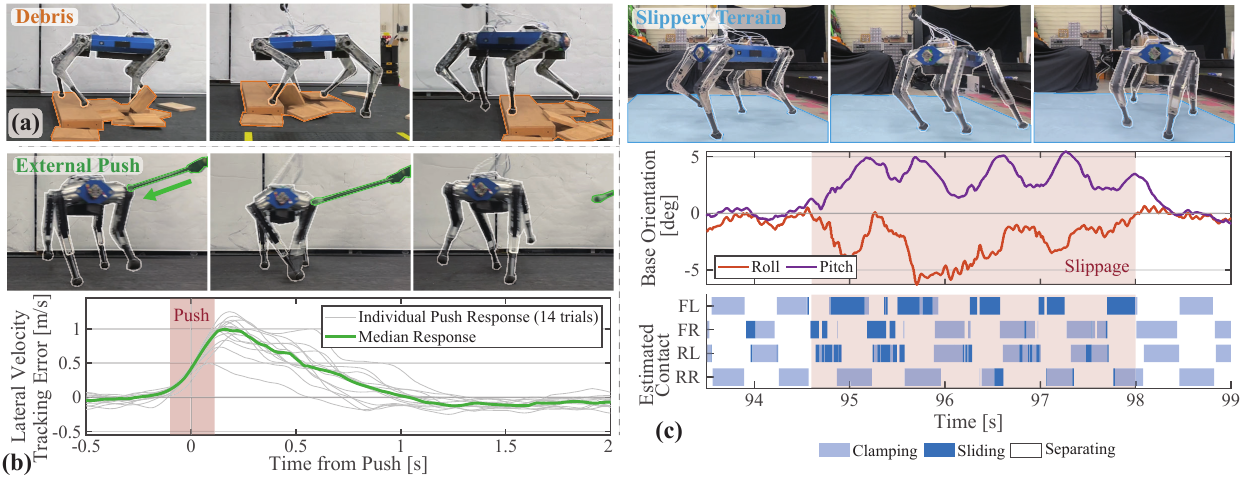}
\caption{Hardware robustness tests of trotting policies trained with 100,000 samples in under 2~min. Snapshots show representative trials for each experiment. (a) Traversal over scattered debris. (b) Recovery from external pushes. The plot shows lateral-velocity tracking error over 14 trials; gray curves show individual trials, the green curve shows the median, and the shaded interval marks the push. (c) Locomotion on a low-friction surface. Base roll and pitch and estimated foot-contact modes are shown during pronounced foot slippage; the highlighted interval marks the slippage event.}
\label{fig:bptt_exp}
\end{figure*}

\subsection{Training across contact gradients}
\label{sec:apg_grad}
Fig.~\ref{fig:apg_main} compares three contact sensitivities under a fixed training setup.  The baselines are the mode-dependent exact sensitivity~\cite{LeLidec2025} (Section~\ref{sec:bg_exact}) and the Signorini-relaxed sensitivity~\cite[Eq.~19]{Kim2023}, which replaces the Signorini complementarity $\lambda_N\sigma_N=0$ with $\lambda_N\sigma_N=\rho>0$ while retaining the unrelaxed friction cone. Appendix~\ref{app:signorini} derives this sensitivity in our formulation. The proposed gradient relaxes the full frictional contact condition. Both relaxed gradients use $\rho=1$. The proposed gradient uses $\delta=0.2$ for trotting and pacing and $\delta=0.5$ for the gait-free tasks. Training uses 16 environments and 5,000 policy iterations per run across 10 seeds.

Four velocity-command-tracking tasks are considered: gait-free locomotion at nominal ($\mu\in[0.6,1.0]$) and low ($\mu\in[0.2,0.4]$) friction, and trotting and pacing at nominal friction. Training command ranges are $[0,1.0]$\,m/s in forward velocity for gait-free tasks and $[-1.0,1.0]$ and $[-0.6,0.6]$\,m/s in forward and lateral velocities for trotting and pacing, respectively. All tasks share a yaw-rate range of $[-0.8,0.8]$\,rad/s. Each final policy is evaluated on 225 test cases, covering 45 commands at five friction coefficients within the training ranges. Evaluation costs are averaged over steps until a fall or the end of each rollout, and the costs are pooled across test cases and seeds in Fig.~\ref{fig:apg_main}(b). A fall is recorded if base height drops below 0.2\,m, base tilt exceeds $45^\circ$, or per-step cost exceeds a prescribed threshold. Fig.~\ref{fig:apg_main}(c) reports fall rates over an extended friction range, including values outside the training ranges.

Fig.~\ref{fig:apg_main} shows that only the proposed gradient learns both gait-free tasks. The learned policies achieve low evaluation costs and maintain low fall rates over the tested friction range. They advance the feet by swinging or sliding, both admissible because the objective contains no foot-clearance term. The exact and Signorini-relaxed gradients do not acquire gait-free locomotion; their training rewards plateau at similar levels, with high evaluation costs and falls in nearly all rollouts. In their rollouts, the stance feet fail to advance with the body, causing the base to pitch forward and fall under forward commands. Neither baseline provides tangential contact-velocity variation at clamped contacts; normal variation from the Signorini relaxation is insufficient for these tasks.

For the gait-free task trained at low friction, the forward rollout encounters sliding more frequently. The sliding-mode sensitivity provides tangential variation, but both baseline policies still fail to acquire locomotion. This result indicates that relying on the forward rollout to encounter useful contact modes while differentiating only within the realized mode does not suffice. The proposed gradient instead provides sensitivity across contact-mode transitions, enabling gait-free locomotion.

For the gait-conditioned tasks, the proposed gradient learns both trot and pace with the lowest median evaluation costs (Fig.~\ref{fig:apg_main}(b)). The exact gradient shows partial task learning, but its policies still have higher fall rates than those trained with either relaxed gradient (Fig.~\ref{fig:apg_main}(c)). The Signorini-relaxed gradient learns both gaits. The prescribed phase and clearance objective specify when each foot should lift, and the Signorini relaxation supplies normal sensitivity for this liftoff, supporting gait learning. However, the Signorini-relaxed sensitivity provides no direction toward the clamping-to-sliding transition at a stance contact and therefore does not reflect the margin of the contact impulse to the friction-cone boundary. Below the training friction range, the Signorini-relaxed policies fall more frequently on slippery terrain, whereas the proposed policies maintain lower fall rates.

\begin{figure*}[t]
\centering
\includegraphics[width=\textwidth]{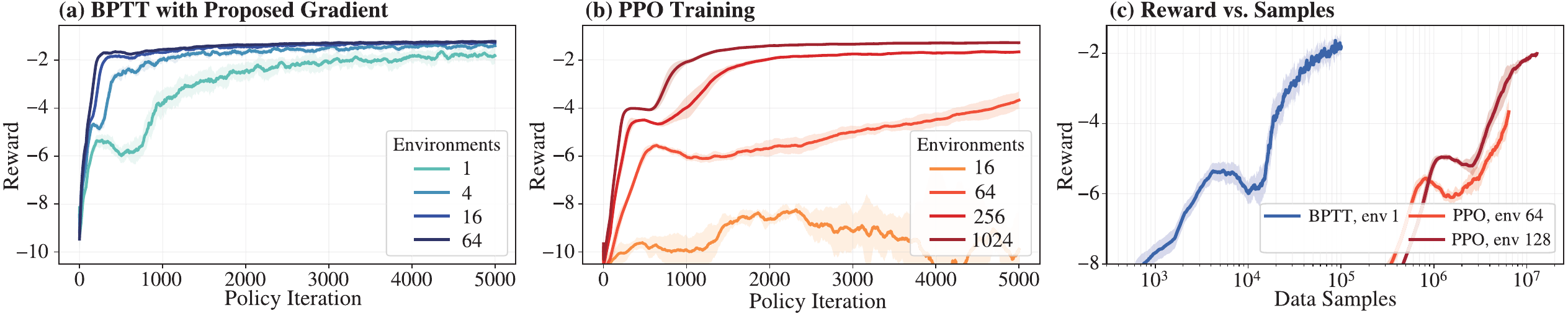}
\caption{Training-reward comparison between BPTT and PPO on the trot command-tracking task. (a) BPTT and (b) PPO training reward over policy iterations for different numbers of environments; curves show the mean and shaded bands denote one standard deviation. (c) Training reward versus cumulative environment steps for BPTT with one environment and PPO with 64 and 128 environments. Among the tested configurations, BPTT learns the trot with one environment, whereas PPO first learns it with 128 environments; PPO with 64 environments does not learn the trot.}
\label{fig:ppo_train}
\end{figure*}

\begin{figure}[t]
\centering
\includegraphics[width=\columnwidth]{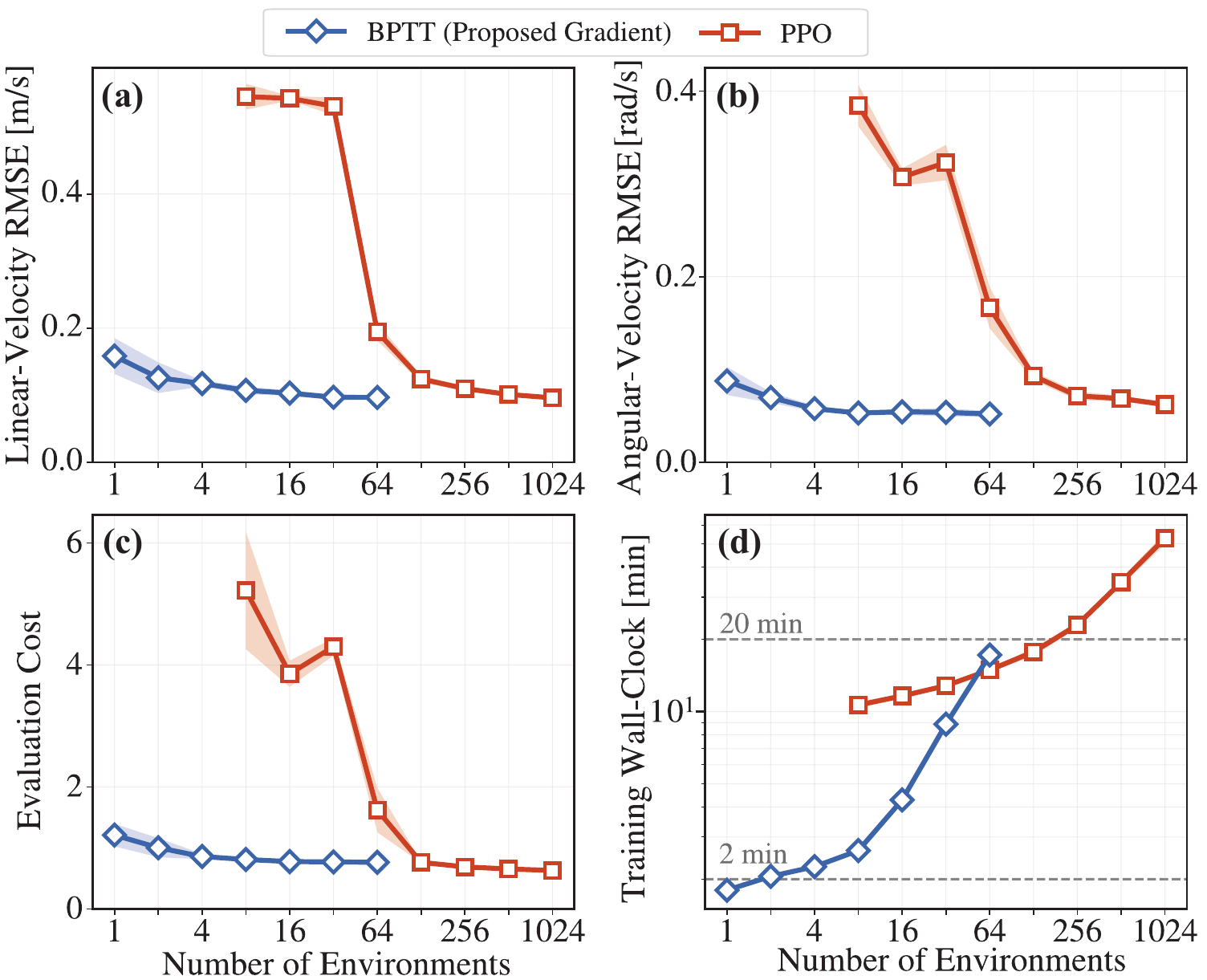}
\caption{Comparison of BPTT and PPO in trained-policy evaluation and training wall-clock time versus the number of environments (mean across seeds; shading denotes one standard deviation). Policy evaluation uses $10$-s trot rollouts over sampled velocity commands and in-distribution friction values. BPTT is shown through 64 environments and PPO through 1024. (a) Linear-velocity tracking RMSE. (b) Angular-velocity tracking RMSE. (c) Evaluation cost. (d) Training wall-clock time on a desktop (log scale); dashed lines mark 2 and 20 minutes.}
\label{fig:ppo_eval}
\end{figure}

\subsection{Hardware demonstration}\label{sec:apg_hw}
Hardware experiments deploy policies learned by BPTT with the proposed gradient on the 45-kg quadruped robot. All policies are trained using a single simulation environment, so each policy update uses one rollout gradient without variance reduction from averaging across parallel environments. The trotting and pacing policies are trained for 5,000 policy iterations with 20 control steps per iteration (100,000 samples), taking 108~s of training time on the desktop platform in Section~\ref{sec:impl}. The low-friction gait-free policy is trained for 8,000 iterations (160,000 samples) and takes 172~s. Fig.~\ref{fig:bptt_exp_gait} shows hardware trials and the corresponding estimated contact patterns for the three locomotion policies. The trotting and pacing policies track commanded velocities on a treadmill while maintaining their prescribed contact patterns. The low-friction gait-free policy is deployed and produces locomotion without a prescribed gait. These policies are transferred to hardware without fine-tuning.

Fig.~\ref{fig:bptt_exp} evaluates the robustness of the trotting policies under debris, external pushes, and low friction. The robot traverses scattered debris while maintaining stable locomotion. Across 14 external-push trials, the lateral-velocity tracking error increases during the push and then decreases toward zero (Fig.~\ref{fig:bptt_exp}(b)). For the low-friction test, a trotting policy trained over the extended range $\mu\in[0.4,1.0]$ is evaluated on a boric-acid-coated surface with measured foot--surface friction coefficient $\mu\approx0.25$, outside the training range. The robot recovers from intermittent foot slips and continues walking without falling. Fig.~\ref{fig:bptt_exp}(c) highlights a pronounced slippage interval. The front-right (FR) and rear-left (RL) feet initially enter sliding, followed by sustained sliding of the front-left (FL) foot. As this slippage persists, the base roll and pitch undergo larger excursions. The attitude excursions then decrease as the robot recovers from the slip and continues locomotion. These hardware results show that the per-rollout gradients provided by the proposed contact sensitivity are sufficient for BPTT policy learning, direct hardware transfer, and robust locomotion under external disturbances. 

\subsection{Comparison with Model-Free RL}\label{sec:apg_rl}
We compare BPTT using the proposed gradient with PPO~\cite{Schulman2017} to assess training sample efficiency and the performance of the resulting policies on the trot task. Fig.~\ref{fig:ppo_train} presents training progress across environment counts, whereas Fig.~\ref{fig:ppo_eval} reports evaluation performance and wall-clock training time. Both methods use the same observation and action spaces, PD control law, RaiSim simulation~\cite{Hwangbo2018}, and per-step cost; its negative, scaled value defines the PPO reward. Each policy iteration collects 20 control steps per environment. BPTT differentiates the resulting undiscounted truncated cost without a terminal value or critic, whereas PPO uses a learned critic to bootstrap the discounted return. Evaluation follows Section~\ref{sec:apg_grad} over the same 225-cell grid, with both velocity commands and friction coefficients within the training distribution. Each configuration uses 5 seeds; the one-environment BPTT case uses 20 seeds. The PPO configuration is summarized in Appendix~\ref{app:ppo}.

BPTT with the proposed gradient learns the trot at every environment count from one to 64 (Fig.~\ref{fig:ppo_train}(a)). PPO does not reach the trot over this range, with unstable training at 16 environments and a reward that increases but converges to a lower value at 64 (Fig.~\ref{fig:ppo_train}(b)). PPO estimates the policy gradient by sampling returns under a stochastic policy, and the variance of this estimate decreases with the batch size of each update~\cite{Recht2019}. The stability of training and the reward attained depend on the batch size, and the trot is recovered only at a sufficiently large batch (Fig.~\ref{fig:ppo_train}(b)). For BPTT with the proposed gradient, the update quality depends on the contact sensitivity rather than on the batch size, and a single environment is sufficient to learn the trot. The
smallest setting that learns the trot uses one environment for BPTT and 128 for PPO. Training uses approximately $10^5$ and $10^7$ samples, respectively, differing by two orders of magnitude (Fig.~\ref{fig:ppo_train}(c)). For PPO, the onset of trotting occurs at 128 environments, where the foot-clearance term defined relative to the reference gait decreases from 0.57 at 64 environments to 0.06 and remains near 0.05 for larger batches.

For velocity tracking and evaluation cost, BPTT attains performance comparable to PPO using fewer environments (Fig.~\ref{fig:ppo_eval}(a)--(c)). The BPTT linear-velocity root-mean-square error (RMSE) reaches 0.126~m/s with 2 environments, compared with 0.124~m/s for PPO with 128 environments. At 8--64 environments, the BPTT linear-velocity RMSE is 0.097--0.107~m/s, closely matching the 0.096--0.110~m/s range attained by PPO at 256--1024 environments (Fig.~\ref{fig:ppo_eval}(a)). The BPTT angular-velocity RMSE reaches 0.058~rad/s with 4 environments, below the 0.062--0.093~rad/s range of PPO at 128--1024 environments. At 8--64 environments, the BPTT angular-velocity RMSE remains at 0.052--0.054~rad/s, below this PPO range (Fig.~\ref{fig:ppo_eval}(b)). For evaluation cost, all BPTT configurations shown attain lower values than PPO with 64 environments. From 16 to 64 environments, the BPTT evaluation cost is 0.765--0.776, comparable to 0.763 for PPO with 128 environments (Fig.~\ref{fig:ppo_eval}(c)). 

For the smallest number of environments that acquire the trot, BPTT with 1 environment trains in 1.8~min, whereas PPO with 128 environments requires 17.7~min, corresponding to a 9.8-fold reduction in wall-clock training time for BPTT (Fig.~\ref{fig:ppo_eval}(d)). The BPTT configurations with 4 and 16 environments, which respectively attain tracking accuracy and evaluation cost comparable to PPO with 128 environments, train in 2.3 and 4.3~min, corresponding to 7.7- and 4.1-fold reductions. At larger BPTT batch sizes, wall-clock time scales linearly with the number of environments, reaching 17.2~min at 64 environments. In this regime, the available CPU cores are saturated, and BPTT computes dynamics gradients through each rollout, increasing the per-environment computational load relative to PPO.

\subsection{Numerical Behavior of the Newton Iteration}
\label{sec:bptt_newton}

\begin{figure}[t]
\centering
\includegraphics[width=\columnwidth]{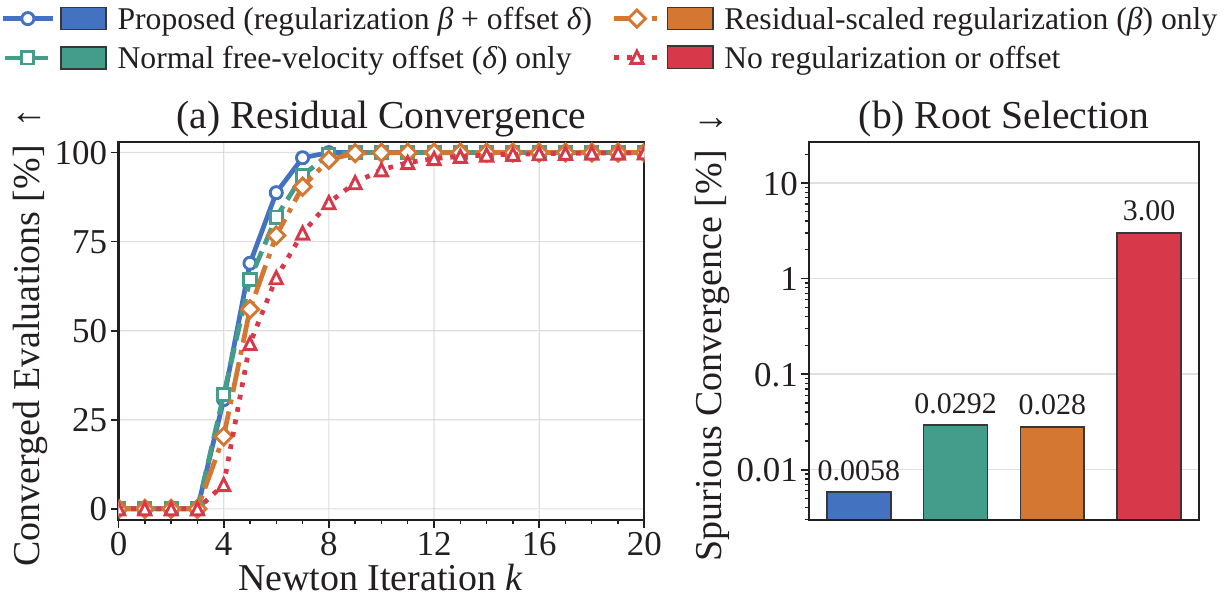}
\caption{Newton convergence and root selection in the backward contact evaluation during BPTT trotting training. (a) Cumulative percentage of all contact evaluations reaching the residual tolerance \(\|F_\rho\|<\varepsilon\) within \(k\) Newton iterations of Algorithm~\ref{alg:newton}. (b) Percentage of all contact evaluations that satisfy the residual tolerance but fail the cone-feasibility check.}
\label{fig:bptt_newton}
\end{figure}

In BPTT trotting training, the proposed configuration satisfies the residual tolerance $\|F_\rho\|<\varepsilon$ ($\varepsilon=10^{-10}$) in $99.95\%$ of backward contact evaluations within eight Newton iterations (Fig.~\ref{fig:bptt_newton}(a)). The ablation independently enables or disables the residual-scaled diagonal regularization $\beta$ and the normal free-velocity offset $\delta$. For each configuration, we collect $5\times10^5$ contact evaluations during single-environment training across five seeds. The curves show the cumulative percentage of evaluations that meet the tolerance within $k$ Newton iterations. Enabling either or both terms yields higher residual-convergence rates during the early iterations than using neither term.

\begin{figure*}[t]
\centering
\includegraphics[width=\textwidth]{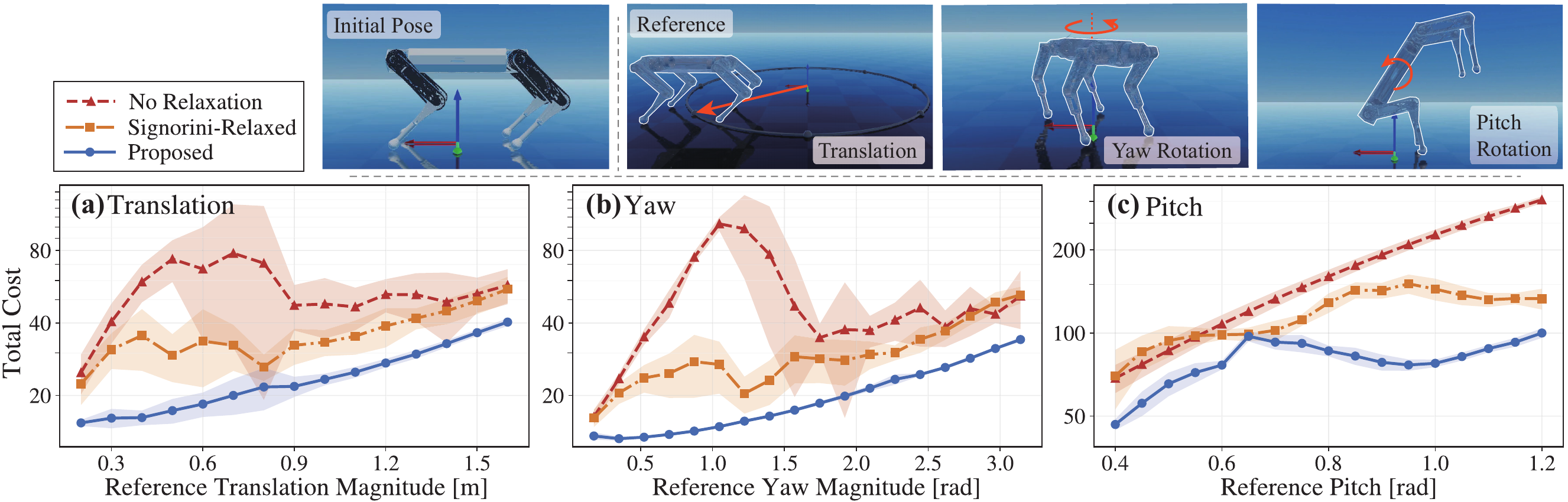}
\caption{Comparison of the three backward contact gradients for CITO base-pose reaching. (a) \textit{Translation}: the reference distance ranges from $0.2$ to $1.6$~m. (b) \textit{Yaw rotation}: the reference yaw ranges from $0.17$ to $3.14$~rad. (c) \textit{Pitch rotation}: the reference pitch ranges from $0.4$ to $1.2$~rad. For each target, the finite-horizon objective is averaged over 400 successive solves along a 10-s trajectory. Lines and shaded regions show the mean and one standard deviation over eight translation directions in (a), ten planar offsets within 0.1~m in (b), and ten planar offsets within 0.3~m in (c). The top row shows the initial standing and reference poses as solid and transparent robots, respectively.}
\label{fig:cito_summary}
\end{figure*}

Residual convergence alone does not guarantee convergence to a physical root because the residual equation $F_\rho(\bm{\lambda})=0$ may also admit nonphysical roots with negative normal impulses, impulses outside the friction cone, or penetrating normal velocities. Spurious convergence denotes residual convergence that fails the feasibility check. If any contact fails this check, we set both transition Jacobians $f_x$ and $f_u$ to zero, thereby blocking gradient propagation through that transition. The spurious-convergence rate over all contact evaluations decreases from $3.00\%$ with neither term enabled to approximately $0.03\%$ with either term alone, and to $0.0058\%$ with both terms enabled (Fig.~\ref{fig:bptt_newton}(b)).
\section{Contact-Implicit Trajectory Optimization}\label{sec:cito}
We compare the proposed backward contact sensitivity with the mode-dependent exact and Signorini-relaxed sensitivities under the same CITO setup. Base-pose tracking tasks compare the optimized trajectories over a range of reference poses. A single-solve comparison examines the optimization iterates from a common initialization and the resulting foot-repositioning decision. An external-pull experiment evaluates the closed-loop robustness of contact-implicit MPC to unmodeled disturbances in simulation.

\subsection{Base-Pose Reaching Without Prescribed Contact Modes}

We evaluate how the contact gradient affects the trajectories and costs obtained by CITO over a range of base-pose references. Fig.~\ref{fig:cito_summary} considers three tasks. (1) \textit{Translation} varies the planar reference displacement from $0.2$ to $1.6$~m over eight directions: forward, backward, left, right, and the four diagonals. (2) \textit{Yaw rotation} varies the base-yaw reference up to $\pi$~rad in both rotation directions, with ten planar positions sampled within $0.1$~m per reference. (3) \textit{Pitch rotation} varies the base-pitch reference from $0.4$ to $1.2$~rad at a $0.1$~m height offset, with ten planar positions sampled within $0.3$~m per reference. All problems start from the nominal standing pose at the origin and are solved under the receding-horizon setting. The Signorini-relaxed and proposed gradients use the relaxation parameter $\rho=1$, and the proposed gradient uses the free-velocity offset $\delta=0.5$. For each target, the cost is the finite-horizon objective averaged over the successive solves.

\begin{figure*}[t]
\centering
\includegraphics[width=\textwidth]{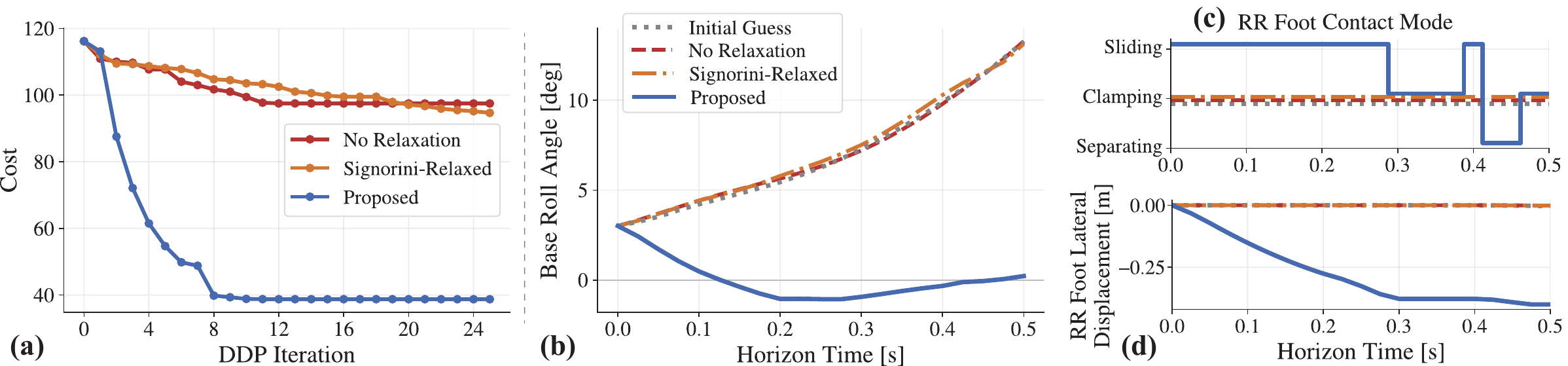}
\caption{Comparison of a single CITO problem over DDP iterations under the three contact gradients. All optimizations use the same current state, base-pose reference, and initial state and control trajectories for the $0.75$-m rightward \textit{Translation} case in Fig.~\ref{fig:cito_summary}(a); the initialization is taken at $t=1.30$~s of the exact-gradient rollout. (a) Cost versus DDP iteration. (b)--(d) Initial guess and optimized trajectories after 25 iterations over the $0.5$-s horizon: (b) base roll angle, (c) RR-foot contact mode, and (d) RR-foot lateral displacement.}
\label{fig:cito_oneshot}
\end{figure*}

Fig.~\ref{fig:cito_summary}(a) and (b) show that the proposed gradient attains the lowest costs throughout the \textit{Translation} and \textit{Yaw rotation} tasks. The proposed gradient provides both normal and tangential contact sensitivity in all contact modes. Consequently, when a foot is clamped in the nominal rollout, the optimizer receives local directions toward separation or sliding, allowing the foot to move from its initially fixed location. In \textit{Translation}, the optimized trajectories reposition the feet as the base moves toward the reference. Similarly, in \textit{Yaw rotation}, they reposition the feet as the body turns, avoiding residual leg twist. As a result, the trajectories reach the references with smaller base-pose and posture residuals. The narrower standard-deviation bands indicate that this foot-repositioning behavior occurs more consistently than with the two baselines.

On the other hand, with the exact gradient, the feet remain clamped at their initial locations for translation references up to $0.7$~m and yaw references up to $1.0$~rad. The base therefore leans toward a translation reference or twists relative to the fixed feet toward a yaw reference, and the cost increases with reference magnitude. At larger references, the greater lean or twist can cause feet to separate or slide during the hard-contact rollout. Although these transitions allow some repositioning and reduce the cost, they do not occur consistently. A foot may reclamp during the rollout, preventing further repositioning. This variation produces the wider standard-deviation bands.

Compared with the exact gradient, the Signorini-relaxed gradient permits normal contact release and lowers the cost over the moderate reference ranges. However, it provides no tangential direction at a clamped contact. During forward \textit{Translation}, feet can therefore remain at their initial or partially relocated positions even as the base approaches the reference, leaving the joints displaced from their nominal configuration. Likewise, during \textit{Yaw rotation}, the base can reach the reference by pivoting about a clamped foot while retaining leg twist. The resulting posture residuals maintain the cost gap to the proposed gradient, while variation in the clamped feet and their locations produces the wider bands.

The proposed gradient also yields the lowest costs across the \textit{Pitch rotation} references (Fig.~\ref{fig:cito_summary}(c)). In this task, reaching the reference requires the rear feet to separate and the robot to balance on the front feet. The proposed contact sensitivity guides both the rear-contact transition and the adjustment of the front-foot locations needed for balance. As a result, the optimized trajectory reduces forward–backward base motion and pitch excursions while tracking the reference. For a representative $1.2$~rad reference, it maintains the pitch within $0.91–1.34$~rad, compared with $0.77–1.50$~rad for the Signorini-relaxed trajectory, and reduces the peak-to-peak forward–backward base displacement from $1.19$ to $0.60$~m.

With the exact gradient, by comparison, all four feet remain clamped. Leg extension raises and pitches the trunk, but the attained pitch remains below the reference and the cost rises as the reference increases. The Signorini-relaxed gradient allows rear-contact separation and thus reduces the cost relative to the exact gradient, consistent with~\cite{Kim2023}. However, its normal-only relaxation does not provide the tangential direction needed to adjust the front feet. The base moves repeatedly forward and backward, alternating between pitch overshoot and undershoot and leaving a cost gap to the proposed gradient.

\begin{figure*}[t]
\centering
\includegraphics[width=\textwidth]{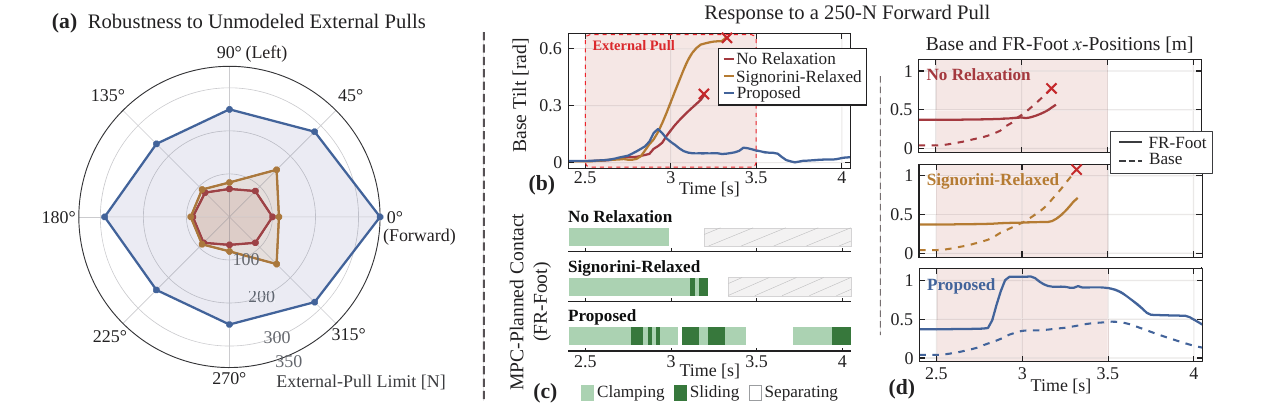}
\caption{Comparison of the three contact gradients in CI-MPC under unmodeled external pulls. (a) Maximum pull magnitude over eight nominal directions. Each value is the largest pull magnitude for which at least $90\%$ of rollouts with perturbed pull directions complete without failure. (b)--(d) Responses to a $250$-N forward pull: (b) base tilt, (c) planned front-right (FR) foot contact mode, and (d) base and FR-foot $x$-positions. Crosses in (d) mark failures due to non-foot body contact with the ground.}
\label{fig:cito_pull}
\end{figure*}

\subsection{Analysis of a Single CITO Problem}\label{sec:cito_single_problem}
We compare how the CITO solution evolves over DDP iterations under the three contact gradients (Fig.~\ref{fig:cito_oneshot}). All three optimizations use identical state and control trajectories for initialization; these trajectories are taken at $t=1.30$~s of the $10$-s rollout obtained with the exact gradient for the $0.75$-m rightward \textit{Translation} case in Fig.~\ref{fig:cito_summary}(a). At this time, the base is moving rightward while the front-right (FR) and rear-right (RR) feet remain clamped. In the full $10$-s rollout, this configuration leads to increasing base roll and eventually a large corrective motion as the base recovers. The shared $0.5$-s CITO initial guess is taken during this clamped phase. Over this $0.5$-s horizon, the planned base roll increases from $3^\circ$ to $13^\circ$ as the base continues moving laterally over the fixed right feet (Fig.~\ref{fig:cito_oneshot}(b)–(d)). 

Starting from the common initialization, the proposed gradient reduces the cost from $116$ to $39$ after $25$ DDP iterations (Fig.~\ref{fig:cito_oneshot}(a)). In the optimized trajectory, both right-foot contacts leave clamping, allowing the front-right (FR) and rear-right (RR) feet to adjust their locations as the base moves rightward. Specifically, the loaded RR contact first transitions from clamping to sliding and moves laterally by approximately $0.4$~m until about $0.29$~s (Fig.~\ref{fig:cito_oneshot}(c)–(d)). It then passes through clamping, sliding, and separating before reclamping at its new location. As a result of this foot repositioning, the planned base roll decreases toward zero (Fig.~\ref{fig:cito_oneshot}(b)).

With the exact and Signorini-relaxed gradients, however, the RR foot remains clamped and its lateral displacement stays at zero. The planned base roll therefore still reaches approximately $13^\circ$, while the costs decrease only to $98$ and $95$, respectively, after $25$ iterations.

\subsection{Push Recovery under Unmodeled Pulls}\label{sec:cito_push}

We evaluate the three contact gradients in contact-implicit model predictive control (CI-MPC) under unmodeled external pulls (Fig.~\ref{fig:cito_pull}). CI-MPC runs at $40$~Hz with a maximum of four DDP iterations per update. The optimized control is executed with a joint-level PD controller, and the simulation is performed in RaiSim~\cite{Hwangbo2018} using a $1$-ms time step and friction coefficient $\mu=1.0$. For this experiment, CI-MPC tracks a nominal standing-pose reference. In each $10$-s rollout, three $1$-s pulls are applied to the base along a fixed body-frame direction, beginning at $2.5$, $5.0$, and $7.5$~s. A rollout is classified as a failure if the optimizer diverges, the base height falls below $0.20$~m, or the base tilt exceeds $45^\circ$. Fig.~\ref{fig:cito_pull}(a) shows the maximum pull magnitude for eight nominal directions. For each direction, the reported value is the largest force magnitude for which at least $90\%$ of rollouts with perturbed pull directions are successful. 

The proposed gradient withstands the largest pulls in every tested direction (Fig.~\ref{fig:cito_pull}(a)), with limits ranging from $240$ to $350$~N and averaging $272$~N. On average, this is $3.4$ times the $81$~N limit obtained with the exact gradient and $2.5$ times the $107$~N limit obtained with the Signorini-relaxed gradient.

For the representative $250$~N forward pull, the optimizer using the proposed gradient transitions the planned front-foot contacts from clamping to sliding (Fig.~\ref{fig:cito_pull}(c)), allowing the feet to advance with the base. As the feet move forward, the resulting impulses oppose the pull, while their repositioning generates a restoring pitch moment and extends support in the pull direction (Fig.~\ref{fig:cito_pull}(d)). Consequently, the base tilt peaks near $10^\circ$ and returns toward the nominal pose between pulses (Fig.~\ref{fig:cito_pull}(b)). The $10$~s rollout completes without failure.

In contrast, with the exact and Signorini-relaxed gradients, the front feet remain clamped at their initial positions as the pull moves the base forward. By approximately $3$~s, the base has moved ahead of the fixed front feet and begins to tip. The body then contacts the ground at $3.19$ and $3.34$~s in the respective rollouts, and both fail (Fig.~\ref{fig:cito_pull}(b)–(d)).

\section{Conclusion}\label{sec:conclusion}
This work proposed DiffContact, a relaxed full-cone contact gradient, for policy learning and contact-implicit trajectory optimization in legged locomotion. The method relaxes the de~Saxc\'e-corrected SOCCP along a central path and computes the relaxed evaluation point using a Newton iteration initialized from the hard-contact solution, while retaining the hard-contact forward rollout. The resulting sensitivity varies smoothly across the contact-mode transitions and recovers the hard-contact fixed-mode sensitivity as $\rho\to0$. Using the proposed gradient, BPTT trained trotting and pacing policies from 100,000 simulator samples each in 108~s. These policies and a low-friction gait-free policy transferred to hardware without fine-tuning; trotting policies were also evaluated in debris traversal, push recovery, and low-friction locomotion. In CITO, the proposed gradient reduced base-pose-reaching costs across translation, yaw, and pitch references and increased the recovered-pull limits relative to the exact and Signorini-relaxed gradients.

Future work will extend the method to non-point feet, uneven terrain, and contacts between arbitrary shapes. These extensions require differentiating through collision detection. The derivation is not specific to legged systems, and its use in contact-rich manipulation will be explored. For longer-horizon policy learning, the present 0.5-s BPTT rollouts could be augmented with a learned terminal value, as in SHAC~\cite{Xu2022}.

\appendices

\section{Jordan Product and Arrow Matrix}\label{app:jordan}
Following~\cite{AlizadehGoldfarb2003,Acary2026}, let
$\bm{w}=(w_0,\bar{\bm{w}})$ and
$\bm{z}=(z_0,\bar{\bm{z}})\in\R\times\R^{d-1}$, where $d=3$
for a point contact. The Jordan product and identity are
\begin{equation*}
\bm{w}\circ\bm{z}
:=
\bigl(\bm{w}^\top\bm{z},\,
w_0\bar{\bm{z}}+z_0\bar{\bm{w}}\bigr),
\qquad
\bm{e}=(1,\bm{0}).
\end{equation*}
The product is represented by the symmetric \emph{arrow matrix}
\begin{equation*}
L(\bm{w})
:=
\begin{pmatrix}
w_0 & \bar{\bm{w}}^\top\\[2pt]
\bar{\bm{w}} & w_0 I_{d-1}
\end{pmatrix},
\qquad
\bm{w}\circ\bm{z}
=
L(\bm{w})\,\bm{z}
=
L(\bm{z})\,\bm{w}.
\end{equation*}

\section{Signorini-Relaxed Contact-Impulse Sensitivity} \label{app:signorini}
The Signorini-relaxed gradient of~\cite{Kim2023} is obtained instead by relaxing
only the normal complementarity. For a normal component with $\lamN>0$,
$\sigN\lamN=\rho$ gives $\sigN=\rho/\lamN$, whose differential is
$d\sigN+(\rho/\lamN^2)d\lamN=0$. Stacking these normal relations with the
unrelaxed tangential relations gives a differential system of the same algebraic
form as~\eqref{eq:compact}, with $\Mlam=\rho D$ and $\Msig=I$, where $D$ is diagonal, 
with entries $1/\lambda[l]^2$ on the normal indices and zero otherwise.
Therefore, the contact-impulse sensitivity with respect to a variable $\eta$ reduces to
\begin{equation*}\label{eq:cimpc}
\frac{\partial\bm{\lambda}}{\partial\eta}
=-\bigl(A+\rho D\bigr)^{-1}
\left(\frac{\partial A}{\partial\eta}\bm{\lambda}
+\frac{\partial\bm{b}}{\partial\eta}\right),
\end{equation*}
which is the Signorini-relaxed contact gradient in~\cite[Eq.~19]{Kim2023}.

\section{PPO Configuration}\label{app:ppo}
Table~\ref{tab:ppo} summarizes the PPO configuration used in Section~\ref{sec:apg_rl}. The environment, observation, action, per-step cost, and episode handling follow the BPTT setup, while PPO bootstraps the return at the end of each 20-step rollout segment using the critic.

\begin{table}[t]
\caption{PPO settings.}
\label{tab:ppo}
\centering\footnotesize
\begin{tabular}{@{}p{0.44\columnwidth}@{}p{0.53\columnwidth}@{}}
\hline
optimizer / learning rate & AdamP / $5\times10^{-4}$\\
discount / GAE $\lambda$ & 0.99 / 0.95\\
clip $\epsilon$ & 0.2\\
epochs / minibatches per update & 10 / 2\\
value / entropy coefficient & 0.5 / 0.01\\
gradient-norm clip & 0.5\\
action std (init / bounds) & 1.0 / $[0.2,\,1.5]$\\
actor / critic nets & $[256,128]$ MLP, LeakyReLU\\
\hline
\end{tabular}
\end{table}

\end{document}